\documentclass[10pt,journal,compsoc]{IEEEtran}
\usepackage{algorithmic}
\usepackage{algorithm}
\usepackage{array}
\usepackage{textcomp}
\usepackage{stfloats}
\usepackage{caption}
\usepackage{url}
\usepackage{verbatim}
\usepackage{graphicx}
\usepackage{cite}
\usepackage{tikz}
\usepackage{comment}
\usepackage{subfigure}
\usepackage{multirow}
\usepackage{makecell}
\usepackage{amsmath,amssymb} 
\usepackage{color}
\usepackage{orcidlink}
\usepackage{booktabs}
\usepackage[accsupp]{axessibility}  
\usepackage{threeparttable}
\usepackage{bbding}
\usepackage{multirow}
\newcommand{\eg}{\emph{e.g.}}
\newcommand{\ie}{\emph{i.e.}}
\DeclareMathOperator*{\argmin}{arg\,min}
\DeclareMathOperator*{\argmax}{arg\,max}
\ifCLASSINFOpdf
\else
\fi
\begin{document}
%
\title{LPT++: Efficient Training on Mixture of Long-tailed Experts}
%
%
%
%

\author{Bowen Dong$^{1,3}$ \quad
Pan Zhou$^{2}$ \quad
Wangmeng Zuo$^{1}$\textsuperscript{\Envelope} \thanks{1. School of Computer Science and Technology, Harbin
Institute of Technology, Harbin 150001, China (e-mail: cswmzuo@gmail.com).
2. School of Computing and Information Systems, Singapore Management University: (e-mail: panzhou3@gmail.com). 
3. The Hong Kong Polytechnic University: (e-mail: bowen.dong@connect.polyu.hk). 
\Envelope \quad denotes corresponding author.}}

%
%

\markboth{Journal of \LaTeX\ Class Files,~Vol.~14, No.~8, August~2015}%
{Shell \MakeLowercase{\textit{et al.}}: Bare Demo of IEEEtran.cls for Computer Society Journals}
%



\IEEEtitleabstractindextext{%
\begin{abstract}

We introduce LPT++, a comprehensive framework for long-tailed classification that combines parameter-efficient fine-tuning (PEFT) with a learnable model ensemble. LPT++ enhances frozen Vision Transformers (ViTs) through the integration of three core components. The first is a universal long-tailed adaptation module, which aggregates long-tailed prompts and visual adapters to adapt the pretrained model to the target domain, meanwhile improving its discriminative ability. The second is the mixture of long-tailed experts framework with a mixture-of-experts (MoE) scorer, which adaptively calculates reweighting coefficients for confidence scores from both visual-only and visual-language (VL) model experts to generate more accurate predictions. Finally, LPT++ employs a three-phase training framework, wherein each critical module is learned separately, resulting in a stable and effective long-tailed classification training paradigm. 
Besides, we also propose the simple version of LPT++ namely LPT, which only integrates visual-only pretrained ViT and long-tailed prompts to formulate a single model method. LPT can clearly illustrate how long-tailed prompts works meanwhile achieving comparable performance without VL pretrained models. 
Experiments show that, with only $\sim$1\% extra trainable parameters, LPT++ achieves comparable accuracy against all the counterparts.
\end{abstract}

}

\maketitle

\IEEEdisplaynontitleabstractindextext

%
\IEEEpeerreviewmaketitle

\section{Introduction}
\IEEEPARstart{L}ong-tailed learning~\cite{Kang2020Decoupling,zhang2021deep} seeks to optimize neural networks trained on datasets with highly imbalanced class distributions, allowing for accurate recognition of objects from both majority and minority classes. However, learning from long-tailed data~\cite{Kang2020Decoupling,zhang2021deep} presents significant challenges in deep learning era. Models must effectively learn to identify both abundant common objects and diverse, yet rare, objects that frequently appear in real-world scenarios~\cite{van2018inaturalist,zhou2017places,gupta2019lvis}. This imbalance often causes networks to overfit to majority classes while neglecting minority classes.  This is because the disproportionate number of training samples from majority classes results in dominant gradients,  hindering the optimization process necessary for recognizing minority classes~\cite{li2019gradient}.

To mitigate this issue, previous methods have focused on three primary strategies to optimize a network from scratch: \textbf{1)} re-sampling the long-tailed data distribution~\cite{Kang2020Decoupling,Li2022Long,li2021metasaug,ren2020metasoftmax} to achieve class balance within each minibatch  during training, \textbf{2)} re-weighting the training loss~\cite{cui2019cbloss,Li2022Long,menon2021longtail} to assign greater importance to minority classes, and \textbf{3)} employing specially-designed techniques such as decoupled training~\cite{Kang2020Decoupling}, knowledge distillation~\cite{li2021self}, or ensemble learning~\cite{zhou2020bbn,wang2020long}. 
Nevertheless, directly learning generalized feature representation and unbiased classifier is still difficult~\cite{cui2021parametric}, since learning with highly diverse and abundant samples from majority classes makes the network bias on corresponding classes. 
Intuitively, training long-tailed learning models from a pretrained model can provide a generalized and balanced feature representation, which makes easier to learn balanced classifier for both majority and minority classes. 
Therefore, pretrained models~\cite{he2016deep,dosovitskiy2021an} are adopted into long-tailed learning~\cite{cui2021parametric,Li2022Long} to conduct fully fine-tuning via the mentioned three training strategies.  
%

Unfortunately, fully fine-tuning pretrained models for long-tailed learning suffers from three main issues. Firstly, as the rapidly increasing size pretrained models, the GPU computation cost and training time are also increased. Hence adapting such models to long-tailed data necessitates whole model fine-tuning, which incurs significantly higher training costs. 
Secondly, fine-tuning the entire model impairs the generalization ability of the pretrained model. Pretrained models trained on large-scale datasets benefit from exposure to abundant data, enabling strong discriminative abilities across various features. Unfortunately, fine-tuning often diminishes this generalization capability due to overfitting to specific features of long-tailed data, making it difficult to handle domain shifts or out-of-distribution data, which are common in long-tailed learning. Finally, fine-tuning results in substantially different models for different learning tasks, compromising model compatibility and increasing deployment costs.

\begin{figure*}[tb]
	\centering
	\subfigure[{Places-LT}]{
		\includegraphics[width=3in]{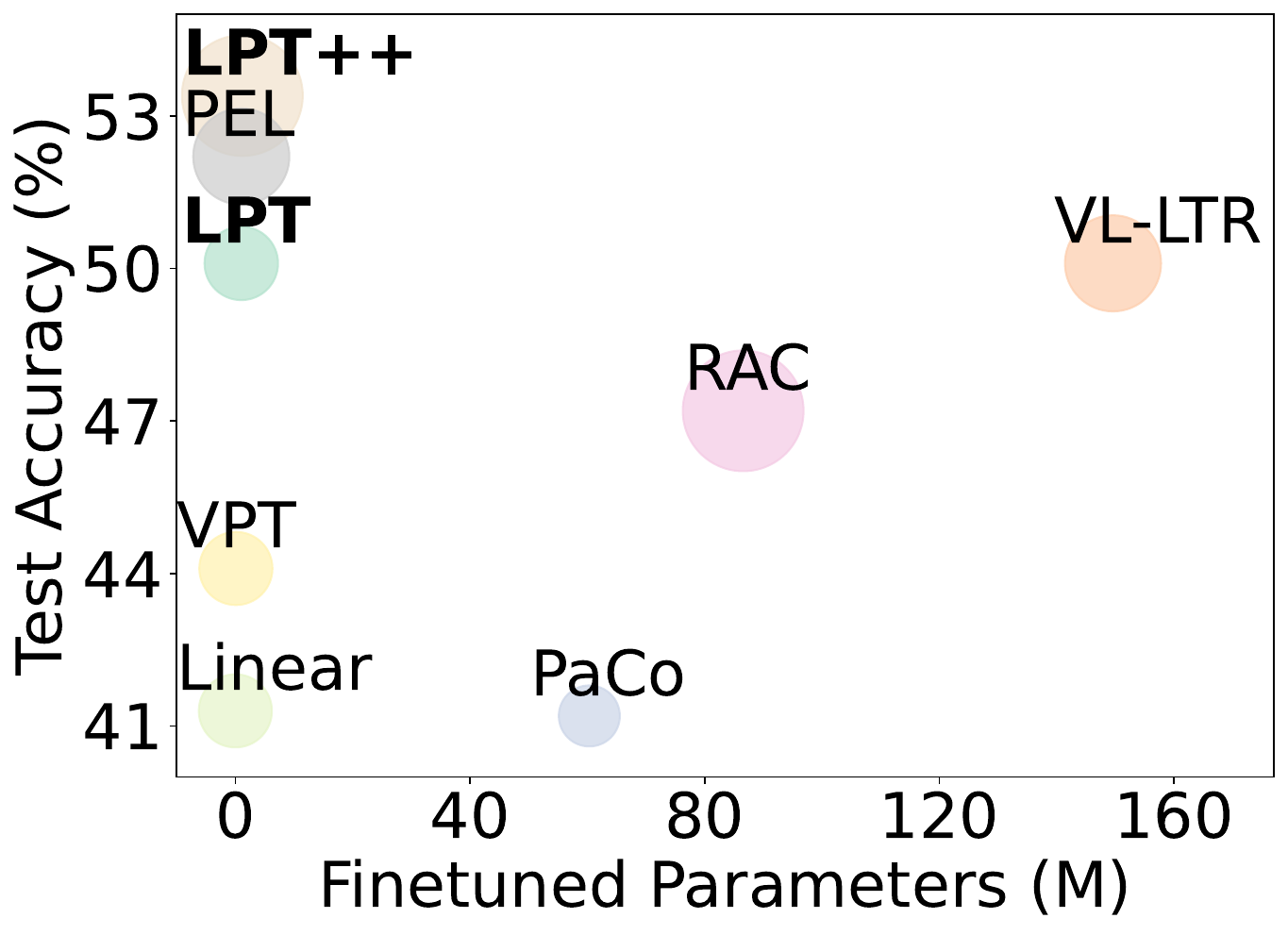}
		\label{fig:per_placeslt}
	} \subfigure[{iNaturalist 2018}]{
		\includegraphics[width=3in]{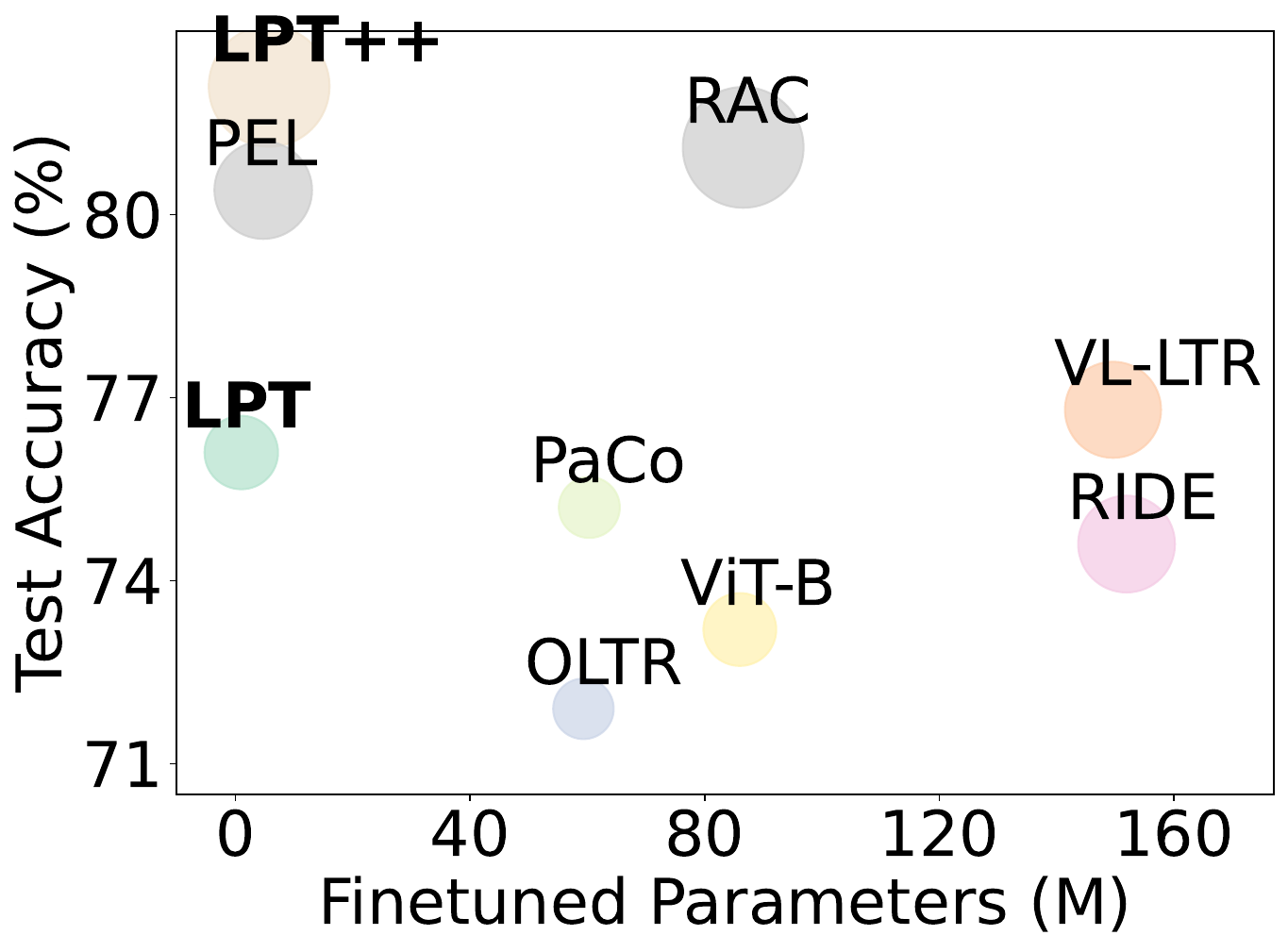}
		\label{fig:per_inat18}
	} \\
	\vspace{-1.1em}
	\caption{Comparison among state-of-the-art long-tailed approaches on Places-LT and iNaturalist 2018, where the size of each spot means the model size of the whole model. LPT++ is our proposed visual-language pretrained long-tailed classification method with mixture of long-tailed experts framework. And LPT is the simple version of LPT++ which removes visual-language pretrained models, visual adapters and mixture-of-experts formulation. 
 LPT and LPT++ only requires $\sim$1\% extra trainable parameters while achieving higher accuracy on two highly long-tailed datasets.  
   }
   \label{fig:param_acc_comp}
	\vspace{-1.3em}
\end{figure*}

\noindent\textbf{Contributions.} 
To address the aforementioned challenges, we propose a novel and effective Long-tailed Prompt Tuning approach   (LPT++), which is a mixture-of-experts (MoE)-enhanced parameter-efficient fine-tuning (PEFT) solution for long-tailed image classification. 
In terms of effective and efficient long-tailed classification, LPT++ incorporates three critical components (\emph{i.e.}, universal long-tailed adaptation modules, mixture of long-tailed experts framework with mixture-of-experts scorer, and multi-phase training framework of LPT++) into pertrained ViTs~\cite{dosovitskiy2021an} for fast adaptation and promising performance.
%
Our contributions are highlighted as follows.

Firstly, we propose universal long-tailed adaptation modules for LPT++, which introduce two types of long-tailed prompts and adapters to learn shared features (knowledge) across all samples and group-specific features for samples with similar characteristics. This approach enhances knowledge learning and the identification of distinct data characteristics. Specifically,   LPT++ utilizes two types of prompts: 1) a shared prompt for all classes, which learns general features and adapts the pretrained model to the target domain, and 2) group-specific prompts that capture features for samples with similar characteristics, improving the model's fine-grained discriminative ability. 
Then LPT++ inserts visual adapters~\cite{chen2022adaptformer}  into ViT blocks to extract discriminative clues among different long-tailed classes. These modules are integrated into a pretrained ViT to formulate a parameter-efficient fine-tuning model.

Secondly, we propose mixture of long-tailed experts framework with corresponding mixture-of-experts (MoE) scorer for LPT++. 
Following the first contribution, one can obtain promising LPT++ single models from visual-only and visual-language pretrained models. Such models can be seen as different model experts for long-tailed classification (\emph{i.e.}, \textbf{long-tailed experts}), and corresponding outputs from the same images can be integrated via ensemble technique~\cite{li2024uni}.  
Therefore, LPT++ leverages generated confidence scores from each model expert as input and calculate a pair of reweighting coefficients for both model experts, which aims to adaptively aggregate such scores into an ensemble result for efficient and precise prediction. 
To achieve the goal of generating reweighting coefficients, inspired by mixture of experts (MoEs)~\cite{li2024uni} which \emph{utilizes additional modules} to route different inputs to different expert models, we propose mixture-of-experts scorer (MoE scorer). With only three-layer MLPs, MoE scorer can efficiently concatenate both confidence scores as input, then predict corresponding reweighting coefficients. MoE scorer is lightweight and can be efficiently and separately optimized, which makes LPT++ both flexible and effective. With calculated coefficients, one can adaptively aggregate confidence scores via weighted averaging. 

Finally, we propose a new three-phase training framework to optimize LPT++ model experts and MoE scorer separately. Such optimization framework can fully exploit the power of each proposed component. In the first phase, LPT++ optimizes the shared prompt, visual adapters, and a classifier   to 1) adapt the pretrained model to the target domain through prompt tuning, and 2) enhance the model's discriminative ability using the trained classifier via adapters, laying the foundation for learning group-specific prompts. In the second phase,  LPT++ trains the group-specific prompts and fine-tunes the classifier from the first phase. For a given input,  LPT++ uses the learned shared prompt to generate a class token, which serves as a query to select matched prompts by computing cosine similarity with the keys from the group-specific prompt set. These matched group-specific prompts, combined with the shared prompts, help the model learn class-specific attributes. Both training phases are performed using the Asymmetric Gaussian Clouded Logit (A-GCL) loss~\cite{dong2023lpt} with a dual sampling strategy. Finally,  the MoE scorer in LPT++  is optimized to adaptively reweight confidence scores from both experts for ensemble learning.

LPT++ effectively addresses three key issues in existing methods. For training cost, LPT++ requires fine-tuning only a small number of prompts, whose size is significantly smaller than the pretrained model, leading to much lower training costs compared to fine-tuning the entire model. Regarding  generalization ability,  LPT++ fine-tunes parameter-efficient modules while keeping the pretrained model frozen, thereby preserving the strong generalization capacity of the original model. Finally, as for compatibility,  LPT++ utilizes specific pretrained models for different tasks, requiring only the storage of small-sized additional parameters, which enhances model compatibility and reduces deployment costs.

Additionally, we also propose Long-tailed Prompt Tuning (LPT) as a simplified version of LPT++, which focuses on visual-only pretrained models and long-tailed prompts. Unlike LPT++, which includes visual-language models, visual adapters and the mixture of long-tailed experts framework, LPT uses a single-model approach, relying on shared and group-specific prompt tuning to optimize only long-tailed prompts with corresponding classifier for higher accuracy. The purpose of LPT is to enable fair comparisons with previous visual-only methods and to evaluate the effectiveness of each prompt type in improving domain adaptation.


As shown in Fig.~\ref{fig:param_acc_comp}, with only ~1\% additional trainable parameters, LPT++ achieves higher accuracy than previous methods that fine-tune the entire pretrained model. Specifically, LPT++ outperforms the state-of-the-art PEL~\cite{shi2023longtail} by 1.2\% and 1.4\% in terms of accuracy on Place-LT~\cite{zhou2017places} and iNaturalist 2018~\cite{van2018inaturalist}, respectively. Further experimental results demonstrate the superiority of LPT++ and its generalization on both long-tailed and domain-shifted data.

\textit{Comparison with our previous conference work.} 
Compared to the ICLR 2023 version of LPT~\cite{dong2023lpt}, the journal version of LPT++ is largely enhanced in terms of both network architecture and training paradigm. Firstly, LPT++ introduces multiple universal adaptation modules rather than prompt-only counterparts for long-tailed classification. Secondly, LPT++ proposes mixture of long-tailed experts framework rather than using single model to improve prediction accuracy. And finally, a new multi-phase training framework is adopted to optimize the whole network efficiently.
Additionally, we also conduct extra quantitative analysis of LPT++ expert models, which includes the effect of different pretrained models, the hyper-parameters of group-specific prompts, and the effect of training strategy. All the comprehensive analysis provides deeper insights into the functionality and efficiency of LPT++.

\section{Related Work}
\subsection{Long-tailed Image Classification}

To address the negative effects of highly imbalanced data distributions, previous works have primarily focused on three aspects: data re-sampling, loss re-weighting, and decoupled training strategies. Data re-sampling methods~\cite{Kang2020Decoupling,li2021metasaug,ren2020metasoftmax} aim to balance the training data between head and tail classes using hand-crafted samplers~\cite{Kang2020Decoupling}, data augmentation techniques~\cite{li2021metasaug}, or meta-learning-based samplers~\cite{ren2020metasoftmax}. Loss re-weighting approaches~\cite{cui2019cbloss,menon2021longtail,Li2022Long} introduce bias into the confidence scores~\cite{menon2021longtail,Li2022Long}, rescale logits using hand-crafted weights~\cite{cui2019cbloss}, or employ meta-learning techniques~\cite{Jamal_2020_CVPR}. Decoupled training strategies and ensemble learning methods~\cite{Kang2020Decoupling,li2021self,zhou2020bbn,wang2020long} further enhance performance on imbalanced datasets. Recently, vision-language-based methods~\cite{ma2021simple,tian2021vl,Long2022} have been proposed, introducing additional language data~\cite{ma2021simple,tian2021vl} or external databases~\cite{Long2022} to generate auxiliary confidence scores, and subsequently fine-tuning the entire CLIP-based model on long-tailed data. Unlike methods that fully fine-tune all parameters, we aim to leverage the powerful, unbiased representation of pretrained models and construct a efficient tuning method to derive a classifier from long-tailed data.



\subsection{Parameter-Efficient Fine-tuning}

Parameter-efficient fine-tuning (PEFT) methods, including prompt tuning~\cite{lester2021power,jia2022vpt}, adapters~\cite{pmlr-v97-houlsby19a,he2022towards}, LoRA~\cite{hu2022lora}, are designed to leverage the representation abilities of pretrained models while fine-tuning only a few trainable parameters to enhance performance on downstream tasks~\cite{zhai2019largescale,zhou2017scene}. In this paper, we focus on prompt tuning~\cite{jia2022vpt}. Specifically, Jia \emph{et al.}~\cite{jia2022vpt} introduced prompt tuning into ImageNet~\cite{deng2009imagenet} pretrained Vision Transformers (ViT)~\cite{dosovitskiy2021an} and optimized the prompts. Wang \emph{et al.}~\cite{wang2022learning} incorporated prompt tuning into a continual learning framework, using multiple learnable prompts to handle various tasks. Distinct from these works, LPT and LPT++ explore the transferability of parameter-efficient fine-tuning with highly imbalanced training data, achieving comparable accuracy and efficiency.
\section{Preliminary Study}\label{sec:preliminary}

\begin{table*}[t]
	\caption{Prompt tuning results on Places-LT~\cite{zhou2017places}. Prompt tuning performs better on overall accuracy  and  few-shot accuracy (\ie~``Few'' in the table)  with different training settings.}
	\vspace{-1em}
	\begin{center}
		\setlength{\tabcolsep}{10.6pt} 
		\renewcommand{\arraystretch}{2.2}
		{ \fontsize{8.3}{3}\selectfont{
				\begin{tabular}{l|c|c|cccc}
					\toprule
					{\bf Method}  &{\makecell{\bf Balanced \\ \bf Sampling}} &{\makecell{\bf Tuned Params\\ \bf (w/o classifier)}} & {\bf Overall} & {\bf Many} & {\bf Medium} & {\bf Few}\\ 
					\midrule
					Linear & - & 0 & 33.29 & 46.48 & 29.45 & 18.77 \\
					VPT & - & 92K & \textbf{37.52} & \textbf{50.42} & \textbf{32.78} & \textbf{23.29} \\
					\midrule
					Linear & \checkmark & 0 & 41.33 & \textbf{49.47} & 41.31 & 27.51 \\
					VPT & \checkmark & 92K & \textbf{44.17} & 45.79 & \textbf{46.73} & \textbf{36.18} \\
					\bottomrule
		\end{tabular}}}
	\end{center}
	\label{table:preliminary}
	\vspace{-2em}
\end{table*}

\subsection{Performance Investigation of VPT}\label{sec:pre_exp}
Previous studies on prompt tuning~\cite{zhou2022learning,jia2022vpt} have focused on fine-tuning with limited data from balanced distributions, leaving its transfer learning capabilities on large-scale long-tailed data~\cite{zhou2017places,van2018inaturalist} unexplored. To initiate our method, we quantitatively evaluate whether prompt tuning benefits long-tailed learning. Specifically, we investigate ViT-B~\cite{dosovitskiy2021an} pretrained on ImageNet-21k~\cite{deng2009imagenet} by comparing the performance of linear probing and a prompt tuning method VPT~\cite{jia2022vpt}, on the large-scale Places-LT dataset~\cite{zhou2017places}. Linear probing fine-tunes a linear classifier on top of a pretrained and fixed feature extractor (\eg, ViT~\cite{dosovitskiy2021an}), whereas VPT concatenates input tokens with learnable prompts (tokens) and a linear classifier atop a pretrained model. During training, we optimize the learnable parameters of these two methods independently for 20 epochs using well-tuned hyperparameters, \eg, SGD with learning rate of 0.02 and weight decay of 1e-4.


Table~\ref{table:preliminary} presents the quantitative results of linear probing and VPT. Without class-balanced sampling, VPT achieves an overall accuracy of 37.52\%, surpassing linear probing by 3.94\%, 3.33\%, and 4.52\% in many-shot, medium-shot, and few-shot accuracy, respectively. Notably, after introducing class-balanced sampling~\cite{Kang2020Decoupling}—which involves randomly sampling classes from the training set and then randomly sampling inputs with equal numbers in each iteration—VPT attains an overall accuracy of 44.17\% and exceeds the counterpart by 8.67\% in few-shot accuracy. Based on these observations, we conclude that: \textbf{a)} prompt tuning consistently enhances overall performance in long-tailed classification, and \textbf{b)} prompt tuning is more robust to long-tailed distributions and provides greater benefits to tail categories. However, from Table~\ref{table:preliminary}, it is evident that the accuracy of prompt tuning on long-tailed problems is insufficient and lags behind state-of-the-art methods.

\begin{figure}[tp]
   \centering
   \begin{minipage}{0.46\textwidth}
      \centering
            
      \includegraphics[width=0.8\textwidth]{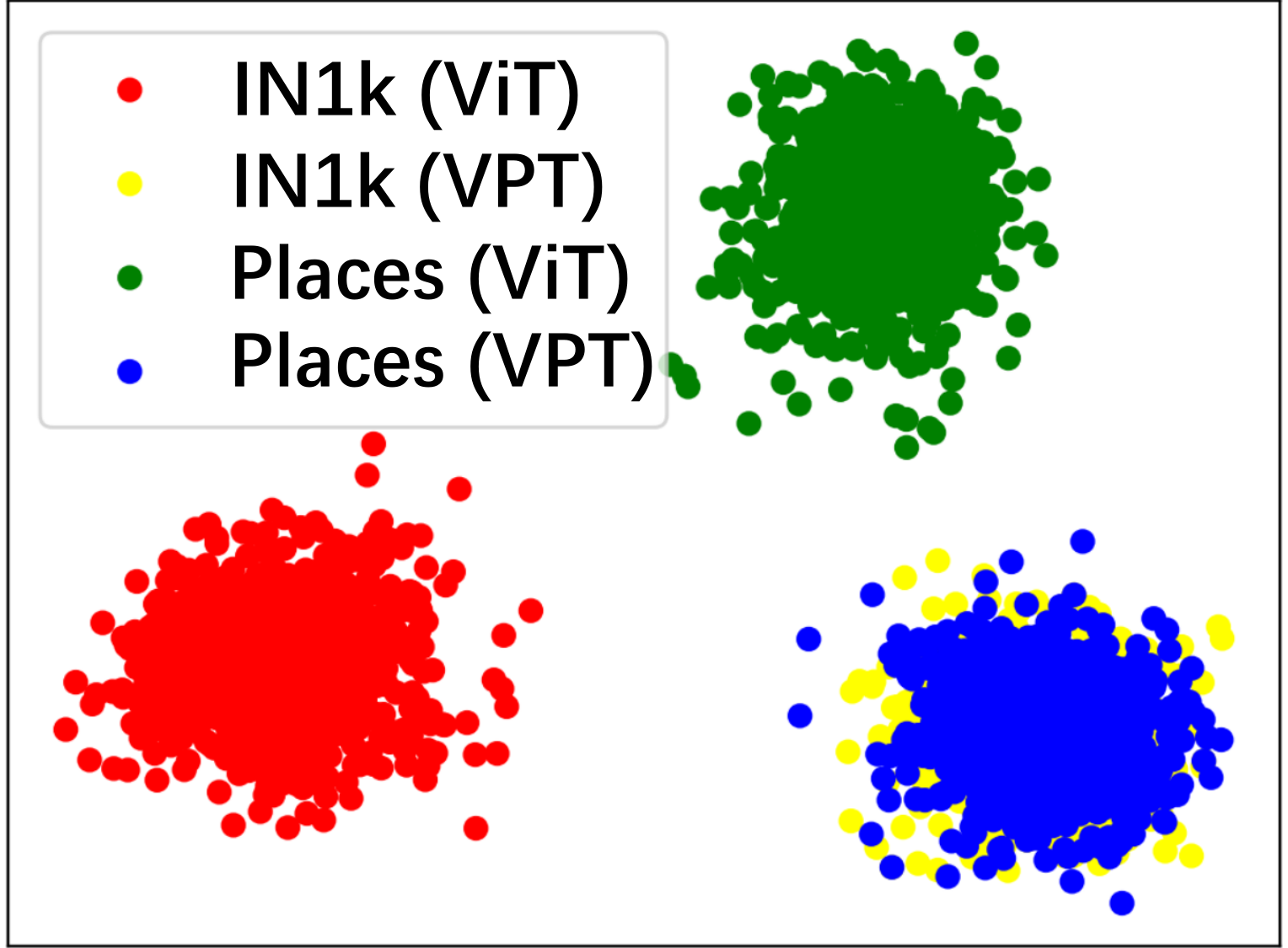}
      \caption{LDA visualization of VPT. } 
      \label{fig:pre_domain_shift}
   \end{minipage}
\hspace{3mm}
   \begin{minipage}{0.5\textwidth}
      \renewcommand\arraystretch{1.3}
      \centering
      \captionsetup{type=table}
      \caption{Analysis of features from ViT-B and VPT.}
    		\setlength{\tabcolsep}{6.6pt} 
  \renewcommand{\arraystretch}{3.5}
      	  		{ \fontsize{8.3}{3}\selectfont{
      \begin{tabular}{l|cc}
         \toprule
         \textbf{Method} & {ViT-B}  & {VPT} \\
         \midrule
         \bf Pretrain Data & IN21k & IN21k \\
         \hline
         \bf Fine-tuned & - & \checkmark \\
         \hline
         \bf Inner-class distance $R_{\text{i}}$ & 2.36$\pm$0.52 & \textbf{1.82$\pm$0.43} \\
         \bf Inner-class / inter-class $\gamma$ & 0.171 & \textbf{0.128} \\
         \bf K-NN Acc & {30.80} & \textbf{31.90} \\
         \bottomrule
      \end{tabular}}}
      \label{table:pre_discriminative}
   \end{minipage}
   \vspace{-2em}
\end{figure}

\subsection{Analysis of Prompt Tuning}\label{sec:pre_analysis}

The reasons behind the improved performance of prompt tuning in long-tailed learning tasks remain unclear. To analyze prompt tuning both quantitatively and qualitatively, we conducted a series of experiments on the Places-LT dataset~\cite{zhou2017places}. We first employed Linear Discriminant Analysis (LDA)~\cite{balakrishnama1998linear} to investigate the learned prompts from a domain adaptation perspective. 
Compared to PCA~\cite{mackiewicz1993principal} and t-SNE~\cite{hinton2002stochastic} which separate samples via unsupervised learning methods, LDA can leverage dataset labels to effectively reduce the data dimension and discriminate the decision boundary of data from different domains.
Specifically, we used the pretrained ViT-B and the ViT-B fine-tuned by VPT on Places-LT (as described in Sec.~\ref{sec:pre_exp}) to extract features from the ImageNet \textit{val} set and Places-LT \textit{val} set. We then used these features to obtain the corresponding LDA vectors for visualization. 
The qualitative results in Fig.~\ref{fig:pre_domain_shift} reveal that: \textbf{a)} for the pretrained ViT-B, the extracted features from ImageNet (red cluster) are far from the features from Places-LT (green cluster); \textbf{b)} for the VPT fine-tuned ViT-B, the extracted features from ImageNet (yellow cluster) align closely with the features from Places-LT (blue cluster). These observations indicate that \textit{the learned prompts in VPT help align the fine-tuned data distribution (Places-LT) with the pretrained data distribution (ImageNet), thereby enabling the pretrained model to adapt to the target domain for long-tailed learning tasks}.


Next we investigate the learned prompt from group-specific perspective. Specifically, for each class in Places-LT, we treat samples  in this class as a group (cluster); then for each group $\text{i}$ ($1\leq \text{i} \leq \text{C}$ with total $\text{C}$ classes in dataset), we calculate average  distance between each sample and its corresponding group center, and views this average  distance as inner-class distance $R_{\text{i}}$ of each group. Furthermore, we also define the inter-class distance $D$ as the average distance between any two group centers, and then calculate the ratio $\gamma$ between the average of inner-class distance $R_{\text{i}}$  and the inter-class distance $D$, namely,  $\gamma = \frac{1}{CD}\sum_{\text{i}}R_{\text{i}}$. Intuitively, for a group, the smaller   inner-class distance $R_{\text{i}}$, the more compact of the group. So if $\gamma$ is smaller, then the groups are more discriminable.   Thus, we use $\gamma$ as a metric to measure whether the learnt features are distinguishable, and report the  statistic results  in Table~\ref{table:pre_discriminative}.  One can observe that features from VPT fine-tuned pretrained model achieves smaller  average  inner-class distance  and also smaller ratio $\gamma$ than those in the vanilla pretrained model,   indicating  that features of different classes in VPT  are easier to be distinguished. Moreover, we also conduct K-NN evaluation between the pretrained  ViT-B and VPT fine-tuned pretrained ViT-B. Table~\ref{table:pre_discriminative} shows that VPT surpasses vanilla pretrained ViT-B by 1.1\% in terms of K-NN accuracy, indicating the higher  discriminative ability of a  VPT  fine-tuned  model.  Therefore, one can conclude that \textbf{2)} \textit{the learned prompt can further improve the discriminative ability of pretrained models, thus benefiting to long-tailed classification problems}.

\begin{figure*}
\begin{center}
\includegraphics[width=0.98\textwidth]{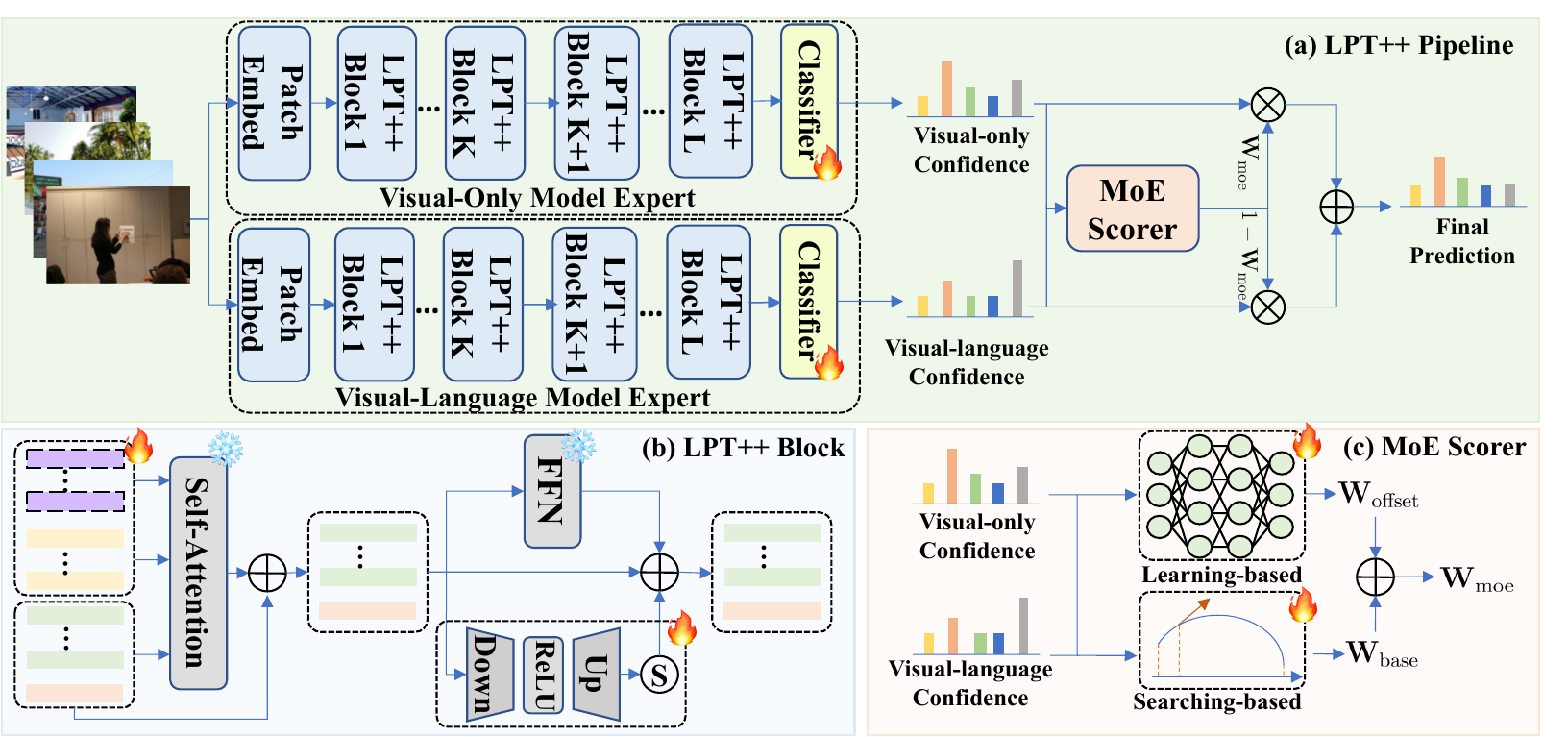}
\end{center}
\vspace{-0.5em}
\caption{\textbf{(a)} Pipeline of LPT++, where \textbf{\textcolor{blue}{snow}} means frozen parameters and \textbf{\textcolor{orange}{fire}} means trainable parameters. LPT++ generates confidence scores via both visual-only and visual-language models, then utilizes MoE scorer to calculate reweighting coefficient $\mathbf{W}_{\text{moe}}$ for final prediction. 
\textbf{(b)} Structure of LPT++ block, where ``S'' means scale operation. 
\textbf{(c)} is the pipeline of MoE scorer, which leverages searching-based scorer to solve shared weight $\mathbf{W}_{\text{base}}$, then uses learning-based scorer to calculate offset $\mathbf{W}_{\text{offset}}$. 
}
\vspace{-1.5em}
\label{fig:pipeline_lptplusplus}
\end{figure*}

Note that naive using class-balanced sampling~\cite{Kang2020Decoupling} or instance-balanced sampling~\cite{Kang2020Decoupling} may lead to severe overfitting on tail classes or head classes~\cite{zhang2021deep} respectively.
To balance accuracy between head classes and tail classes and avoid overfitting, we propose dual sampling strategy. Specifically, for each training iteration in Phase 2, LPT randomly samples a mini-batch $\{\mathbf{I}\}_{\text{ins}}$ from instance-balanced sampler as well as another mini-batch $\{\mathbf{I}\}_{\text{bal}}$ from class-balanced sampler. For samples in $\{\mathbf{I}\}_{\text{bal}}$, we simply set $\beta=1$ to calculate $\mathcal{L}_{\text{P}_{\text{2}}}$; and for samples in $\{\mathbf{I}\}_{\text{ins}}$, we set $\beta=\eta(\text{E}-\text{e})/\text{E}$, where $\eta$ is the initialized weight for $\{\mathbf{I}\}_{\text{ins}}$, $\text{E}$ denotes the maximum number of epochs, and $\text{e}$ is the current epoch number.  


\section{LPT++: Mixture of Long-tailed Experts}\label{sec:method}
\subsection{Overview}

The findings from Sec.~\ref{sec:preliminary} motivate us to develop an efficient and effective long-tailed learning approach centered on prompt tuning~\cite{jia2022vpt,tian2021vl,Long2022,dong2023lpt,shi2023longtail}. To this end,  based on parameter-efficient fine-tuning (PEFT), 
we propose a novel framework called LPT++  for long-tailed classification. As illustrated in Figure~\ref{fig:pipeline_lptplusplus}, LPT++ consists of three key components. 
Firstly, we propose ``\emph{Universal Long-tailed Adaptation Module}'', which integrates both long-tailed prompts~\cite{dong2023lpt,jia2022vpt} and visual adapters~\cite{pmlr-v97-houlsby19a} into long-tailed learning. Specifically, LPT++ operates by incorporating a shared prompt and multiple visual adapters, which are designed for all classes to capture general features and knowledge. This facilitates the adaptation of a pretrained model to the target domain while simultaneously enhancing its discriminative capabilities on the training data. Furthermore, LPT++ employs group-specific prompts to extract features unique to each group, thereby refining the classifier used in the initial phase for improved performance. The optimization of these prompts is carried out through shared prompt tuning and group-specific prompt tuning, forming the core of the long-tailed prompt tuning (LPT) pipeline. Enhanced by class-centric initialization and various foundation models (visual-only pretrained and visual-language pretrained), this multi-phase approach allows for the training of both visual-only and visual-language LPT++. After efficient adaptation, the pretrained visual-only and visual-language models with proposed universal long-tailed adaptation module can be seen as model experts for corresponding task. 
Secondly, we propose ``\emph{Mixture of Long-tailed Experts}'' framework as well as corresponding ``\emph{Mixture-of-Expert (MoE) Scorer}''. With confidence scores of specific images from optimized LPT++ model experts, MoE scorer feds confidence scores into a lightweight MLP to calculate weight coefficients. These coefficients are leveraged to reweight confidence scores from visual-only and visual-language model experts to obtain a more precise prediction. 
And finally, we formulate a new multi-phase long-tailed training methodology specifically designed for LPT++. Rather than naive joint training for all new modules, we decompose the training of universal adaptation modules and MoE scorer, thus formulating a three-phase training procedure. The decomposed training framework with corresponding designed training objectives ensures that each module can be effectively and efficiently optimized separately, thus improving the stability of training framework as well as the final accuracy. 
In the following sections, we will introduce the critical components of LPT++ in detail.

\subsection{Universal Long-tailed Adaptation Module}
Our first contribution is the proposed universal long-tailed adaptation module, which incorporates both long-tailed prompts and visual adapters for better long-tailed accuracy.
Specifically, based on the observation in Sec.~\ref{sec:pre_analysis}, we first aim to introduce shared prompt to make pretrained ViTs to adapt to domain long-tailed data and improve the discriminative ability in the whole long-tailed domain. 
Utilizing a pretrained ViT~\cite{dosovitskiy2021an} with $\text{L}$ layers, our objective is to optimize the shared prompt $\textbf{u} = \left[\textbf{u}_{\text{1}},\dots,\textbf{u}_{\text{L}}\right]$ and the cosine classifier $f(\cdot;\theta_{f})$. 
Here, $\textbf{u}$ adheres to the structure defined by VPT-Deep~\cite{jia2022vpt}, comprising $\text{L}$ individual learnable token sequences. 
Specifically, given an input image $\textbf{I}$, LPT++ initially derives the patch tokens $\mathbf{z}_{\text{0}}$ using the pretrained patch embedding layer. Subsequently, employing the pretrained transformer encoder and a given class token (\text{[CLS]}) $\mathbf{c}_{\text{0}}$. For the $i$-th layer in ViT, where $1\leq \text{i} \leq \text{L}$, the query used in the $i$-th block is defined as $\mathbf{q}^{\text{attn}}_{\text{i}}=\left[\mathbf{c}_{\text{i-1}},\mathbf{z}_{\text{i-1}}\right]$, with corresponding key and value $\mathbf{k}^{\text{attn}}_{\text{i}}=\mathbf{v}^{\text{attn}}_{\text{i}}=\left[\mathbf{c}_{\text{i-1}},\mathbf{z}_{\text{i-1}},\mathbf{u}_{\text{i}}\right]$. Then we update $(\mathbf{c}_{\text{i}},\mathbf{z}_{\text{i}})$ with $\mathbf{u}$:
\begin{equation}\label{eqn:phase1block}
   (\mathbf{c}_{\text{i}},\mathbf{z}_{\text{i}}) = \text{FFN}_{\text{i}}(\text{Attn}_{\text{i}}(\mathbf{q}^{\text{attn}}_{\text{i}}, \mathbf{k}^{\text{attn}}_{\text{i}}, \mathbf{v}^{\text{attn}}_{\text{i}})),
\end{equation}
where $[\cdot, \ldots, \cdot]$ denotes a token concatenation operation along the token number direction,  $\text{Attn}_{\text{i}}$ and $\text{FFN}_{\text{i}}$ represent the self-attention layer and feed-forward network in the $i$-th pretrained ViT block~\cite{vaswani2017attention}, respectively. The final class token $\mathbf{c}_{\text{L}}$ is then fed into the cosine classifier $f$ to calculate per-class confidence scores  $\textbf{s} = f(\mathbf{c}_{\text{L}}; \theta_{f})$. 
Meanwhile, He \emph{et al.}~\cite{he2022towards} have shown that adapters~\cite{chen2022adaptformer} can enhance the discriminative capability of models across many-shot and few-shot classes by introducing non-linear low-rank features in the feed-forward network (FFN). Therefore, instead of solely fine-tuning shared prompts, we also insert Adaptformers~\cite{chen2022adaptformer} into the FFN layers of transformer blocks. This approach enables parameter-efficient fine-tuning of both prompts and visual adapters. By employing multiple parameter-efficient modules concurrently, LPT++ facilitates accurate domain adaptation and enhances the discriminative representation, thereby achieving higher accuracy in long-tailed learning tasks.

In addition to using shared prompt to adapt target long-tailed data to ease the learning difficulty, we also aim to mitigating the long-tailed learning issue via parameter-efficient fine-tuning modules. A straightforward approach to mitigating the challenges of long-tailed learning involves dividing the training data into multiple groups based on feature similarity. This allows for the sharing of group-specific knowledge within each group, thereby reducing recognition difficulty. Motivated by this, we aim to utilize different group prompts to manage samples from various classes, facilitating the collection of group-specific features and enhancing the pretrained model's fine-grained discriminative ability. Consequently, we introduce group-specific prompts, each comprising $\text{m}$ individual learnable prompts
$\mathcal{R} = \{(\mathbf{k}_{1}, \mathbf{r}^{\text{1}}), \dots , (\mathbf{k}_{\text{m}}, \mathbf{r}^{\text{m}})\}$, where $\mathbf{k}_{\text{i}}$ is the key of the corresponding i-th group prompt $\mathbf{r}^{\text{i}}$ and each $\mathbf{r}^{\text{i}}$ has $\text{L}-\text{K}$ trainable token sequences. 
To reduce computational cost and the number of additional parameters, we use only the shared prompt in the first $\text{K}$ blocks and introduce the group-specific prompt set $\mathcal{R}$ into the last $\text{L}-\text{K}$ blocks. 

\subsubsection{Class-Centric Initialization. }
LPT~\cite{dong2023lpt} utilizes random initialization to set up the weights of the classification head, a method that introduces no prior knowledge of class definitions and may consequently constrain the final performance quality. To address this limitation, we propose a universal initialization approach known as class-centric initialization (CC-Init). Specifically, for visual-only model expert of LPT++ (\emph{i.e.}, LPT++(V)),  the centroid of $i$-th class $\phi^{\textbf{vo}}_{i}$ is defined as the mean feature of training samples belonging to the $i$-th class:
\begin{equation}
    \phi^{\textbf{vo}}_{i} = \frac{1}{n^{\text{train}}_{i}}\sum_{\mathbf{I}\in \mathbb{C}_{i}}g_{\text{vo}}(\mathbf{I}),
\end{equation}
where $n^{\text{train}}_i$ indicates the number of training samples from $i$-th class, $\mathbb{C}_{i}$ means the $i$-th class name, and $g_{\text{vo}}$ represents the frozen visual-only pretrained model used in LPT++. Then, we use the calculated $\phi^{\textbf{vo}}$ to initialize $\theta^{\text{vo}}_{f}$, which is the parameter of visual-only cosine classifier $f^{\text{vo}}$.

And for the visual-language model expert of LPT++ (\emph{i.e.}, LPT++(VL)), which benefits from aligning with extensive image-text pairs, one can adopt the zero-shot classification methods~\cite{radford2021learning,zhou2022learning} to initialize using text prompts (\eg, ``a photo of [classname]'') for initialization. However, such simple text prompts may lack the detailed class definitions necessary to enhance the class discrimination ability of LPT++(VL). Instead, we propose LLM-driven class-centric initialization. Specifically, for the $i$-th class, we first use the reliable class definition $T^{\text{seed}}_{i}$ crawled by VL-LTR~\cite{tian2021vl} as seed definition. Then, leveraging state-of-the-art LLMs (\eg, GPT-4~\cite{openai2023gpt4}), we generate a concise yet precise definition by:
\begin{equation}
T^{\text{llm}}_{i} = \text{LLM}(\text{[TASK]:[}T^{\text{seed}}_{i}\text{]}),    
\end{equation}
where [TASK] represents ``summarize the definition of [$\mathbb{C}_{i}$]'' and $\mathbb{C}_{i}$ means the $i$-th class name. After extracting the centric of $i$-th class by $\phi^{\text{vl}}_{i} = g_{\text{text}}(T^{\text{llm}}_{i})$, where $g_{\text{text}}$ means the CLIP text encoder, one can utilize $\phi^{\text{vl}}$ to initialize the weight of visual-language cosine classifier $\theta^{\text{vl}}_{f}$.

\subsection{Mixture of Long-tailed Experts (MoLEs) Framework}
\subsubsection{Pipeline of MoLEs}
The primary contribution of LPT++ lies in our proposed mixture of long-tailed experts (MoLEs) framework. Illustrated in Figure~\ref{fig:pipeline_lptplusplus}(a), this framework comprises two essential components: the visual-only and visual-language base models of LPT++, and a pivotal lightweight mixture-of-experts scorer (MoE scorer). 
Specifically, with given visual-only and visual-language model experts, one can compute the visual-only confidence scores $\mathbf{\hat{s}}_{\text{vo}} = g_{\text{vo}}\cdot f_{\text{vo}}(\mathbf{I})$ and the visual-language counterpart $\mathbf{\hat{s}}_{\text{vl}} = g_{\text{vl}}\cdot f_{\text{vl}}(\mathbf{I})$, where $f_{\text{vo}}$ and $f_{\text{vl}}$ indicate the backbone of LPT++ base models, $g_{\text{vo}}$ and $g_{\text{vl}}$ mean corresponding classifiers. To adaptively reweight and fuse the final predition score $\mathbf{\hat{s}}_{\text{moe}}$, we leverage a lightweight MoE scorer $\psi$ to calculate the reweighting coefficient by $\mathbf{W}_{\text{moe}} = \psi(\mathbf{\hat{s}}_{\text{vo}}, \mathbf{\hat{s}}_{\text{vl}})$. Finally, we calculate the final confidence score $\mathbf{\hat{s}}_{\text{moe}}$ as follows:
\begin{equation}\label{eq:s_moe}
    \mathbf{\hat{s}}_{\text{moe}} = \mathbf{W}_{\text{moe}}\mathbf{\hat{s}}_{\text{vo}} + (1-\mathbf{W}_{\text{moe}})\mathbf{\hat{s}}_{\text{vl}}.
\end{equation}
In the following, we will describe the architecture of the MoE scorer and its inference pipeline. And the corresponding training details will be stated in Sec.~\ref{sec:training}.

\subsubsection{Lightweight Mixture-of-Experts Scorer}
Next, we outline the design of our mixture-of-experts scorer (MoE scorer) in LPT++. MoE scorer $\psi$ comprises two main components: a searching-based scorer $\psi_{s}$ and a learning-based scorer $\psi_{l}$.
Both scorer require the visual-only confidence scores $\mathbf{\hat{s}}_{\text{vo}}$ and the visual-language counterpart $\mathbf{\hat{s}}_{\text{vl}}$. 
For searching-based scorer $\psi_{s}$, the linear combination of $\mathbf{\hat{s}}_{\text{vo}}$ and $\mathbf{\hat{s}}_{\text{vl}}$ is a convex problem, thus one can leverage a binary search method to find a sub-optimal weight $\mathbf{W}_{\text{base}}$ by:
\begin{equation}
    \mathbf{W}_{\text{base}} = \argmin_{\mathbf{W}}[\sum_{i=1}^{n^{\text{train}}}(\argmax_{\mathbf{c}}\mathbf{W}\mathbf{\hat{s}}_{\text{vo}}^{i}+(1-\mathbf{W})\mathbf{\hat{s}}_{\text{vl}}^{i} = \mathbf{y}_{i})]
\end{equation}
Using $\mathbf{W}_{\text{base}}$ to reweight the confidence scores has been shown to improve accuracy. However, to achieve adaptive adjustment of weights for each input sample and further enhance accuracy, we propose a learning-based MoE scorer $\psi_{l}$. $\psi_{l}$ employs a lightweight 3-layer MLP $\psi_{l}(\mathbf{\hat{s}}_{\text{vo}}, \mathbf{\hat{s}}_{\text{vl}})$ with hidden channel of 2048 to predict the weight offset $\mathbf{W}_{\text{offset}}$. Specifically, with given input confidences $\mathbf{\hat{s}}_{\text{vo}}$ and $\mathbf{\hat{s}}_{\text{vl}}$, the scorer concatenates these scores into a vector, projects it into a hidden embedding of dimension 2048, and finally outputs a scalar representing $\mathbf{W}_{\text{offset}} = \psi_{l}(\mathbf{\hat{s}}_{\text{vo}}, \mathbf{\hat{s}}_{\text{vl}})$.
Therefore, the final reweighting coefficient $(\mathbf{W}_{\text{moe}}, 1-\mathbf{W}_{\text{moe}})$ from our MoE scorer are calculated by $\mathbf{W}_{\text{moe}} = \mathbf{W}_{\text{base}} + \mathbf{W}_{\text{offset}}$.
Hence the final confidence score $\mathbf{\hat{s}}_{\text{moe}}$ is obtained by Eq.~\ref{eq:s_moe}.

Employing both searching-based and learning-based scorers in our MoE scorer offers two distinct advantages. Firstly, relying solely on a learning-based scorer to directly predict weights could potentially complicate optimization. Conversely, learning the offset of weights can facilitate training by providing a more manageable adjustment process. Secondly, both components are automatically constructed without human intervention or additional data, ensuring the MoE scorer's ability to generalize effectively across diverse target scenarios.


\begin{figure*}
\begin{center}
\includegraphics[width=0.98\textwidth]{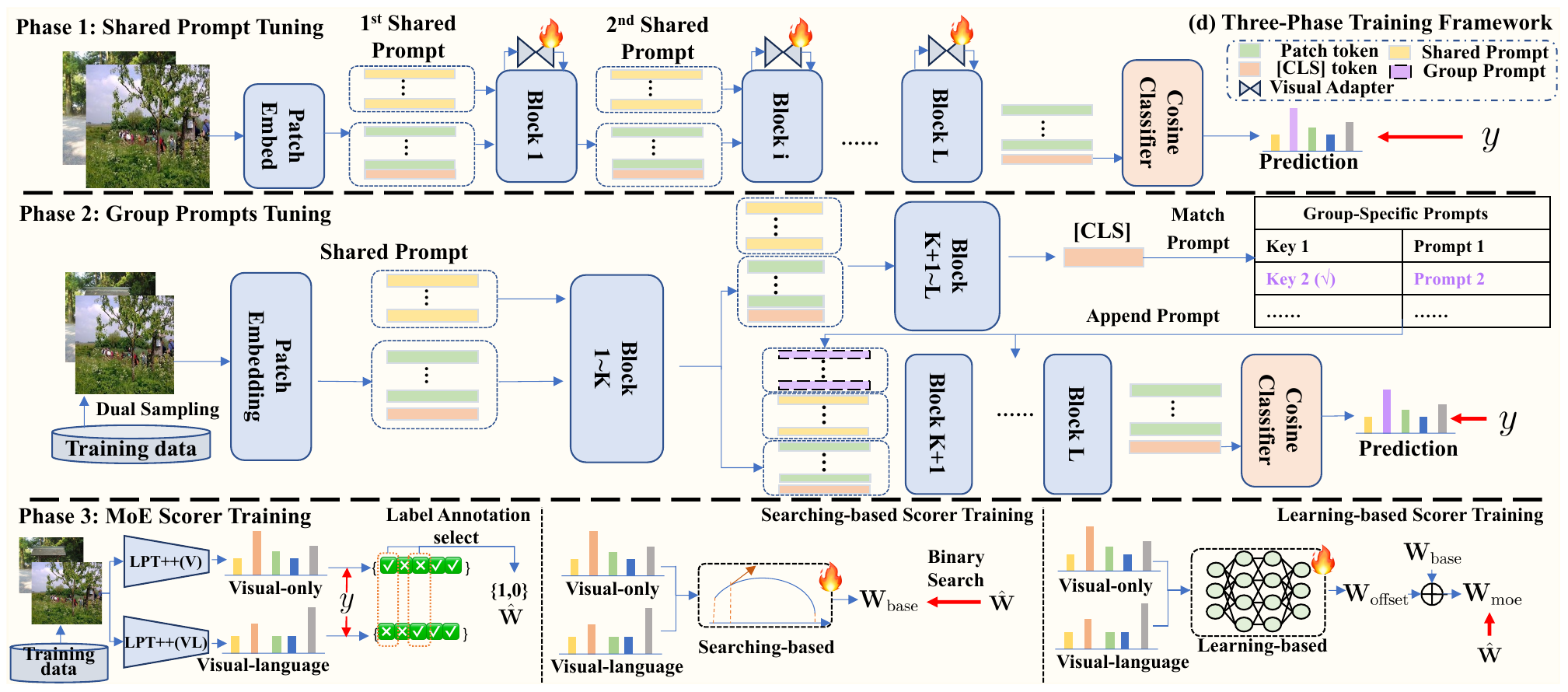}
\end{center}
\vspace{-0.5em}
\caption{Multi-phase training framework of LPT++. In phase 1, LPT++ optimizes both shared prompt and visual adapters simultaneously to adapt pretrained model to target domain and improve the discriminative ability. In phase 2, LPT++ freezes the learned shared prompts and visual adapters, and optimizes the group-specific prompts to further improve the discriminative ability. Both phases bring the visual-only and visual-language LPT++ model experts. And in phase 3, with confidence scores from both model experts, LPT++ optimizes searching-based and learning-based scorer to adaptively reweight confidence scores.
}
\vspace{-1.5em}
\label{fig:pipeline}
\end{figure*}

\subsection{Multi-Phase Training Framework of LPT++}\label{sec:training}
After disussing the architecture, we elucidate the training details of LPT++. 
As shown in Fig.~\ref{fig:pipeline_lptplusplus}(d), to improve the stability of training framework for better performance, we separate the training framework of LPT++ into three phases, \emph{i.e.}, \textbf{shared prompt tuning}, \textbf{group prompt tuning}, and \textbf{MoE scorer training}. In the initial phase, we optimize both the shared prompts and AdaptFormers within both base models to facilitate the adaptation of pretrained models to the desired long-tailed domain. 
In the second phase, we maintain the same training pipeline as in LPT to optimize the group-specific prompts. 
Finally, in a supplementary third phase, we extract confidence scores from both the visual-only and visual-language base models for training images. Subsequently, we optimize the MoE scorer using filtered training data annotated with binary labels automatically.
In the following, we illustrate the training details of LPT++.

\subsubsection{Phase 1: Shared Prompt Tuning}\label{sec:method_phase1}
%
With given ground-truth $\mathbf{y}$ of corresponding input $\mathbf{I}$, we minimize $\mathcal{L}_{\text{P}_{\text{1}}} = \mathcal{L}_{\text{cls}}(\mathbf{s}, \mathbf{y})$ during the training of phase 1 to optimize $\textbf{u}$ and $\theta_{f}$, where $\mathcal{L}_{\text{cls}}$ is the classification loss used in both phases and will be discussed in Sec.~\ref{sec:method_loss}.

\subsubsection{Phase 2: Group Prompts Tuning}\label{sec:method_phase2}
%

In this section,  we focus on the training procedure of group prompts tuning. Specifically, based on our observation \text{(2)} in Sec.~\ref{sec:pre_analysis}, we select the query $\mathbf{q} = \mathbf{c}_{\text{L}}$ from Phase 1 rather than using output class token from pretrained ViT like~\cite{wang2022learning}, since the class token $\mathbf{c}_{\text{L}}$ typically exhibits stronger discriminative ability. Given the query $\mathbf{q}$, we adaptively select the best-matched prompts from $\mathcal{R}$ by $\left[\text{w}_{\text{1}}, \dots , \text{w}_{{k}}\right] = \text{top-k}(\langle \textbf{q}, \left[\textbf{k}_{\text{1}}, \dots, \textbf{k}_{\text{m}}\right]\rangle, {k})$,
where $\text{top-k}(\cdot,{k})$ returns the indices of prompts $\textbf{w}=\left[\text{w}_{\text{1}}, \dots , \text{w}_{{k}}\right]$ with the largest $k$ cosine similarities, and $\langle \cdot,\cdot\rangle$ means the cosine similarity operator. 

Here, we discuss the optimization of keys. A straightforward approach might be to enforce queries from the same class to match specific keys. However, this method is impractical due to the difficulty in determining which classes should match certain prompts precisely. Instead, we opt to minimize the distance between the matched queries and keys, thereby optimizing these keys adaptively. We design the query function from this perspective. As observed in Sec.~\ref{sec:pre_analysis}, the feature cluster of each class generated by the fine-tuned Phase 1 is compact.
Therefore, for queries from the same class, if we randomly select a query $\mathbf{q}_{\text{i}}$ and a key $\textbf{k}^{'}$ then minimize $1-\langle\mathbf{q}_{\text{i}},\mathbf{k}^{'}\rangle$, the distance between $\textbf{k}^{'}$ and other queries are naturally minimized, given that these queries are fixed and sufficiently compact. Therefore, during training, each key is learned to be close to one or multiple nearby clusters, ultimately guiding the corresponding group prompt to gather group-specific features.
%
%

Moreover, since \textbf{1)} VPT~\cite{jia2022vpt} benefits from prompt ensemble, and \textbf{2)} may enhance the recognition of samples from tail classes, LPT++ performs prompt ensemble with multiple selected prompts instead of using only one matched group prompt from $\mathcal{R}$, which is shown as $\mathbf{r} = \text{sum}(\left[\mathbf{r}^{\text{w}_\text{1}}, \dots, \mathbf{r}^{\text{w}_{k}}\right]) / {k}$,
thus resulting an ensemble group prompt $\mathbf{r}$. 
With given $\mathbf{r}$, LPT++ reuses the feature $(\mathbf{c}_{\text{K}},\mathbf{z}_{\text{K}})$ from Phase 1 as $(\mathbf{\hat{c}}_{\text{K}},\mathbf{\hat{z}}_{\text{K}})$ to save computational cost, then define the query used in $\text{i}$-th block as $\mathbf{\hat{q}}^{\text{attn}}_{\text{i}}=\left[\mathbf{\hat{c}}_{\text{i-1}},\mathbf{\hat{z}}_{\text{i-1}}\right]$, and key with value $\mathbf{\hat{k}}^{\text{attn}}_{\text{i}}=\mathbf{\hat{v}}^{\text{attn}}_{\text{i}}=\left[\mathbf{\hat{c}}_{\text{i-1}},\mathbf{\hat{z}}_{\text{i-1}},\mathbf{u}_{\text{i}}, \mathbf{r}_{\text{i-K}}\right]$,
finally update $(\mathbf{\hat{c}}_{\text{i}},\mathbf{\hat{z}}_{\text{i}}) = \text{FFN}_{\text{i}}(\text{Attn}_{\text{i}}(\mathbf{\hat{q}}^{\text{attn}}_{\text{i}}, \mathbf{\hat{k}}^{\text{attn}}_{\text{i}}, \mathbf{\hat{v}}^{\text{attn}}_{\text{i}}))$,
where $\text{K+1}\leq \text{i}\leq \text{L}$ indicates the index of the last $\text{L} - \text{K}$ pretrained blocks in ViT.
Next, the output class token $\mathbf{\hat{c}}_{\text{L}}$ are fed into the cosine classifier $f$ and calculate per-class confidence scores by $\mathbf{\hat{s}} = f(\mathbf{\hat{c}}_{\text{L}}; \theta_{f})$. 
After phase 2, one can utilize different pretrained models~\cite{radford2021learning,dosovitskiy2021an} to obtain corresponding LPT++ model experts.
%

\subsubsection{Phase 3: MoE scorer Training}
Finally, we outline the training of the MoE scorer. For the shared weight term $\mathbf{W}_{\text{base}}$, we leverage a variant of binary search algorithm with a loose error threshold $\epsilon = 1e-3$ to minimize search time while maintaining accuracy. For the learning-based MLP scorer $h$, we gather the training samples from original long-tailed training set which both LPT++ base models output conflict predictions, and automatically annotate the binary ground-truth $\mathbf{y}_{\text{moe}}$ based on the ground-truth class label $\mathbf{y}$, \emph{i.e.}, $\mathbf{y}_{\text{moe}}=1$ means LPT++(V) is correct and vice versa. It's noteworthy that after human verification, we confirm the collected dataset's balance, ensuring the reliability of the learning-based scorer.  Subsequently, we utilize the obtained $\mathbf{y}_{\text{moe}}$ with corresponding $\mathbf{\hat{s}}_{\text{vo}}$ and $\mathbf{\hat{s}}_{\text{vl}}$ to optimize $h$. 

\subsubsection{Loss Function}\label{sec:method_loss}
Finally, we discuss the training losses in each training phase. 
During the first phase, we minimize the classification loss $\mathcal{L}_{\text{cls}}$ to optimize shared prompt, visual adapter, and the classifier simultaneously. 
To further eliminate the negative effect from highly imbalanced data distribution, we propose the asymmetric GCL loss $\mathcal{L}_{\text{A-GCL}}$. This loss adjusts logits based on statistical label frequency from the training data and re-weights the gradient between positive and negative classes. For illustration, we use $\mathbf{\hat{s}} = f(\mathbf{\hat{c}}_{\text{L}}; \theta_{f})$ which is calculated in the Phase 2 of LPT++ as example to demonstrate $\mathcal{L}_{\text{A-GCL}}$. Following Li \emph{et al.}~\cite{Li2022Long}, we re-scale the confidence score of $i$-th class as follows:
\begin{equation}
   \textbf{v}_{\text{i}} = \alpha(\mathbf{\hat{s}}_{\text{i}} - (\log n_{\text{max}} - \log n_{\text{i}})\left\|\epsilon\right\|)
\end{equation}
where $\alpha$ is the scaling factor, $\epsilon$ is the random variable from gaussian distribution, $n_{\text{i}}$ and $n_{\text{max}}$ mean the label frequency of i-th class and the maximum label frequency in the training set, respectively. 
Then, we calculate per-class probability $\mathbf{p}=\left[\textbf{p}_{\text{1}}, \dots, \textbf{p}_{\text{C}}\right]$ by $\left[\textbf{p}_{\text{1}}, \dots, \textbf{p}_{\text{C}}\right] = \text{softmax}(\left[\textbf{v}_{\text{1}}, \dots, \textbf{v}_{\text{C}}\right])$.
Next, we use asymmetric re-weighting~\cite{ridnik2021asymmetric} to eliminate the effect from negative gradient in long-tailed learning.
Suppose $\text{j}$ is ground-truth class of $\textbf{I}$, we define $\mathcal{L}_{\text{A-GCL}}$ as:
\begin{equation}\label{eqn:loss_agcl}
   \mathcal{L}_{\text{A-GCL}} = (1-\textbf{p}_{\text{j}})^{\lambda_{+}}\log(\textbf{p}_{\text{j}}) + \sum\nolimits_{1\leq \text{i}\leq \text{C}, \text{i}\neq \text{j}}(\textbf{p}_{\text{i}})^{\lambda_{-}}\log(\textbf{p}_{\text{i}}),
\end{equation}
where $\lambda_{+}$ and $\lambda_{-}$ is the focusing parameter~\cite{lin2017focal} for ground-truth class and negative classes respectively.
We leverage the asymmetric GCL loss $\mathcal{L}_{\text{A-GCL}}$ with dual sampling proposed in LPT~\cite{dong2023lpt} to optimize visual-only and visual-language model experts. Hence we define the phase 1 loss by $\mathcal{L}_{\text{P}_{\text{1}}} = \mathcal{L}_{\text{cls}}$.

And during the second phase, we aim to train group-specific prompts, hence LPT++ relys on the classification loss to optimize group prompts and a key matching loss $\mathcal{L}_{\text{key}}$ to optimize corresponding keys in prompt group. For classification loss, we follow phase 1 and minimize $\mathcal{L}_{\text{cls}}$. And for $\mathcal{L}_{\text{key}}$, we minimize the cosine similarity between query $\mathbf{q}$ and corresponding matched keys $\left[\mathbf{k}_{\text{w}_{\text{1}}}, \dots, \mathbf{k}_{\text{w}_{{k}}}\right]$:
\begin{equation}
    \mathcal{L}_{\text{key}} = (1 - \frac{1}{{k}}\sum\nolimits_{\text{i}\in \textbf{w}}\langle \mathbf{q}, \mathbf{k}_{\text{i}}\rangle).
\end{equation}
Therefore, the training objective of phase 2 $\mathcal{L}_{\text{P}_{\text{2}}}$ is defined as:
\begin{equation}
    \mathcal{L}_{\text{P}_{\text{2}}} = \beta\mathcal{L}_{\text{cls}}(\mathbf{\hat{s}}, \mathbf{y}) + (1 - \frac{1}{{k}}\sum\nolimits_{\text{i}\in \textbf{w}}\langle \mathbf{q}, \mathbf{k}_{\text{i}}\rangle),
\end{equation}
where $\beta$ is scale factor of $\mathcal{L}_{\text{cls}}$ and will be discussed later.

Finally, for the third phase, we formulate the binary classification problem in MoE scorer as an regression problem, \emph{i.e.}, enforce the coefficient $\mathbf{W}_{\text{moe}}$ to match the annotated coefficient of LPT++(V) $\hat{\mathbf{W}}$. Specifically, $\hat{\mathbf{W}}=1$ means LPT++(V) is correctly predicted, and $\hat{\mathbf{W}}=0$ means LPT++(VL) is correct. Then we adopt MSE loss to minimize both $\mathbf{W}_{\text{moe}}$ and $\hat{\mathbf{W}}$ by $\mathcal{L}_{\text{P}_{\text{3}}} = \text{MSE}(\mathbf{W}_{\text{moe}}, \hat{\mathbf{W}})$.


\subsection{LPT: A Simpler Version of LPT++}
In addition to the full LPT++, we introduce a simplified variant termed \emph{Long-tailed Prompt Tuning (LPT)}. Specifically, LPT incorporates visual-only pretrained models (\emph{i.e.}, ImageNet-21K pretrained ViT~\cite{dosovitskiy2021an}) alongside our proposed long-tailed prompts (\emph{i.e.}, shared prompt and group-specific prompts), and formulate the prompt-based long-tailed classification framework by the same method of LPT++. Compared to LPT++, LPT removes visual-language pretrained models and mixture of long-tailed experts framework, thus formulate a single model based long-tailed classification method. Therefore, LPT only relys on both shared prompt tuning and group prompt tuning phase in Fig.~\ref{fig:pipeline} to optimize the additional prompts as well as classifiers. The reason why we propose LPT is two-fold. Firstly, using visual-only pretrained model to build up LPT can make fair comparison with previous visual-only pretrained state-of-the-art methods. Secondly, LPT only introduces long-tailed prompts into pretrained VIT, which ensures us to clearly investigate effectiveness of each kind of prompt and qualitatively analyze domain adaptation capability. 

 
\begin{table*}[t]
\caption{Comparison with state-of-the-art long-tailed classification methods on Places-LT dataset~\cite{zhou2017places}. 
}
\vspace{-1em}
\label{table:placeslt}
\begin{center}
   \setlength{\tabcolsep}{9.6pt} 
	\renewcommand{\arraystretch}{2.8}
	  		{ \fontsize{8.3}{3}\selectfont{
\begin{tabular}{l|c|c|c|c|cccc}
\toprule
{\bf Method} & {\bf Backbone} &{\makecell{\bf Tuned \\ \bf Params}}&{\makecell{\bf Total \\ \bf Params}} &{\makecell{\bf Extra Data}} & {\bf Overall} & {\bf Many} & {\bf Medium} & {\bf Few}\\ 
\midrule
\multicolumn{6}{l}{\textbf{Visual-only Pretrained}} \\
\midrule
OLTR~\cite{openlongtailrecognition} & Res152 & 60.34M & 60.34M & - & 35.9 & 44.7 & 37.0 & 25.3 \\
LWS~\cite{Kang2020Decoupling} & Res152 & 60.34M & 60.34M & - & 37.6 & 40.6 & 39.1 & 28.6 \\
PaCo~\cite{cui2021parametric} & Res152 & 60.34M & 60.34M & - & 41.2 & 36.1 & 47.9 & 35.3  \\
VPT~\cite{jia2022vpt} & ViT-B &\textbf{0.09M} & 86.66M & - &37.5 &\textbf{50.4} &33.8 &23.3 \\
LPT (Ours) & ViT-B &\textbf{1.01M} & 87.58M & - &\textbf{50.1}& 49.3& \textbf{52.3}& \textbf{46.9} \\
\midrule
\multicolumn{7}{l}{\textbf{Visual-Languge Pretrained}} \\
\midrule
RAC~\cite{Long2022} & ViT-B & 86.57M & 236.19M & IN21k Feat & 47.2 & 48.7 & 48.3 & 41.8 \\
BALLAD~\cite{ma2021simple} & ViT-B & 149.62M & 149.62M & - & 49.5 & 49.3 & \textbf{50.2} & \textbf{48.4} \\
VL-LTR~\cite{tian2021vl} & ViT-B &149.62M & 149.62M & Wiki Text & \textbf{50.1}& \textbf{54.2}& 48.5& 42.0 \\
LPT++ (Ours) & ViT-B &\textbf{1.19M} & 236.19M & - &\textbf{53.4}& 51.9& \textbf{54.9}& \textbf{52.7} \\
\bottomrule
\end{tabular}}}
\vspace{-2em}
\end{center}
\end{table*}
\begin{table}[h]
\centering
\begin{minipage}[t]{0.44\textwidth}
   \centering
   \caption{Comparison results on iNaturalist 2018. }
   \setlength{\tabcolsep}{8.6pt} 
	\renewcommand{\arraystretch}{3.2}
	  		{ \fontsize{8.3}{3}\selectfont{
   \begin{tabular}{l|cccc}
      \toprule
      \textbf{Method} & \textbf{Overall} & \textbf{Many} & \textbf{Medium} & \textbf{Few} \\
      \midrule
      \multicolumn{3}{l}{\textbf{Vision-only Pretrained}} \\
      \midrule
      PaCo & 75.2 & - & - & 74.7 \\
      ViT-B/16 & 73.2 & - & - & - \\
      ViT-L/16 & 75.9 & - & - & - \\
      LPT (Ours) & {76.1} & 62.1& 76.2 & {79.3} \\
      \midrule
      \multicolumn{3}{l}{\textbf{Vision-Language Pretrained}} \\
      \midrule
      VL-LTR & 76.8 &-&- & - \\
      RAC & \underline{80.2} &\textbf{75.9}& \underline{80.5} & \underline{81.0} \\
      PEL & 80.4 & 74.0 & 80.3 & 82.2 \\
      LPT++ (Ours) & \textbf{82.1} & \underline{74.8} & \textbf{82.0} & \textbf{83.9} \\
      \bottomrule
   \end{tabular}}}
   \label{table:comp_inat18}
\end{minipage}
\vspace{-2em}
\end{table}

\section{Experiments}\label{sec:exp}

\subsection{Datasets and Evaluation Protocol}\label{app:datasets}
%

In line with previous works \cite{openlongtailrecognition, Kang2020Decoupling}, we evaluate the performance of LPT++ on two challenging benchmarks, Places-LT and iNaturalist 2018.
We also report results on out-of-distribution ImageNet~\cite{wang2019learning,imageneta,imagenetr,imagenetv2}. More details are in the suppl.

\subsection{Implementation Details}\label{sec:impl_detail}

In LPT++, we use ImageNet-21k pretrained ViT-B/16~\cite{dosovitskiy2021an} and CLIP ViT-B/16~\cite{radford2021learning} as pretrained backbones. The default prompt length is set to 10, with prompts applied across all transformer blocks in the ViT. For group-specific prompts, we fix the number of shared layers at $\text{K} = 6$ and set the prompt size to $\text{m} = 20$.
For visual adapter~\cite{chen2022adaptformer}, the hidden dimension is set to 8 for Place-LT and 256 for iNaturalist 2018 to accommodate the varying number of classes. 
During the first two phases, we use the SGD optimizer with a momentum of 0.9 new modules. 
We employ a cosine learning rate scheduler with an initial learning rate of $0.002 \times \frac{B}{256}$ and 5 warmup epochs, where $B$ denotes the batch size. In the asymmetric GCL loss, we set $\lambda_{+} = 0$ and $\lambda_{-} = 4$. For Phase 2, the initialized weight $\gamma$ used in ${\mathbf{I}}_{\text{ins}}$ is set to 0.5. 
And for phase 3, we optimize the MoE scorer for 50 epochs with the learning rate of 0.01. 
And for LPT, we use similar training framework but without phase 3 to optimize LPT. 
\subsection{Comparison with State-of-The-Art Methods}\label{sec:comp_sota}

\textbf{Results on Places-LT. } 
Methods for long-tailed learning can generally be divided into two categories: visual-only pretrained methods and visual-language (VL) pretrained methods. VL-based approaches~\cite{tian2021vl,Long2022,ma2021simple} often utilize additional data, such as Wiki text or the external ImageNet-21k database, during both training and testing. In contrast, our LPT method which removes the visual-language expert and MoE scorer in LPT++ as mentioned in Sec.~\ref{sec:method} falls into the first category, operating without reliance on extra data. 
As shown in Table~\ref{table:placeslt}, LPT achieves an overall accuracy of 50.1\% and a few-shot accuracy of 46.9\%, with only 1.01M (1.1\%) additional trainable parameters. This performance surpasses the state-of-the-art PaCo~\cite{cui2021parametric} by 8.9\% and 11.6\%, respectively. Even when compared to VL-LTR~\cite{tian2021vl}, a VL-based method that incorporates extra data, LPT matches the overall accuracy while achieving higher few-shot accuracy. Notably, with the integration of our mixture of long-tailed experts and without using additional training data, LPT++ improves overall accuracy by a significant 3.3\% and outperforms all other methods by at least 1.2\%. These results highlight the effectiveness of LPT and LPT++ in handling long-tailed data distributions.


\begin{table*}
   \centering
   \caption{Full comparison with different fine-tuning methods on six different OOD dataset. All methods start from the same IN21K pretrained ViT-B feature extractor. Quantitative results show that LPT++(V) achieves the best accuracy. 
   }
   \vspace{-0.5em}
   \setlength{\tabcolsep}{7pt} 
	\renewcommand{\arraystretch}{3.5}
	  		{ \fontsize{8.3}{3}\selectfont{
   \begin{tabular}{l|c|c|c|c|c|c}
      \toprule
      \textbf{Method} & ImageNet-Sketch & ImageNet-ReaL & ImageNet-V2 & ImageNet-A & ImageNet-R & ObjectNet \\
      \midrule
      Linear Probe & 31.55 & 81.43 & 63.54 & 29.20 & 45.72 & 6.61 \\
      Fully Fine-tune & 32.25 & 80.10 & 62.31 & 30.12 & 43.14 & 7.13 \\
      VPT & 34.64 & 85.82 & 68.51 & 35.17 & 47.06 & 8.03 \\
      WISE-FT & 34.79 & 82.20 & 65.76 & 36.75 & 47.32 & 8.00 \\
      \hline
      LPT++(V) & \textbf{36.22} & \textbf{87.22} & \textbf{70.71} & \textbf{39.65} & \textbf{50.47} & \textbf{8.22} \\
      \bottomrule
   \end{tabular}}}
   \label{table:comp_robustness_full}
\vspace{-1em}
\end{table*}
\begin{table*}[t]
   \caption{Effect of shared prompt tuning and group-specific prompt tuning phase in LPT++ on Places-LT~\cite{zhou2017places}.}
    \vspace{-1em}
   \label{table:ablation}
   \begin{center}
      \setlength{\tabcolsep}{9.6pt} 
	\renewcommand{\arraystretch}{3.2}
	  		{ \fontsize{8.3}{3}\selectfont{
   \begin{tabular}{l|cccc|cccc}
   \toprule
   {\bf Method}  & \bf Prompt & \textbf{Phase 1} & $\mathcal{L}_{\text{A-GCL}}$ & \textbf{Phase 2} & \bf Overall & \bf Many & {\bf Medium} & {\bf Few}\\ 
   \midrule
   Linear & - & - & - & - & 33.29 & 46.48 & 29.45 & 18.77 \\
   VPT & \checkmark & - & - & - & 37.52 & \textbf{50.42} & 33.78 & 23.29 \\
   \midrule
   \text{(a)} & - & \checkmark & - & - & 41.33 & 49.47 & 41.31 & 27.51  \\
   \text{(b)} & \checkmark & \checkmark & - & - & 49.10 & \textbf{49.62} & 51.53 & 43.25 \\
   \text{(c)} & \checkmark & \checkmark & \checkmark & - & \textbf{49.41} & 46.89 & \textbf{52.54} & \textbf{47.32}  \\
   \text{(d)} & \checkmark & \checkmark & \checkmark & \checkmark &\textbf{50.07}& 49.27& \textbf{52.31}& \textbf{46.88} \\
   \bottomrule
   \end{tabular}}}
   \vspace{-1em}
\end{center}
\end{table*}
\begin{table*}[t]
   \caption{Ablation study of each phase in LPT++ on Places-LT benchmark~\cite{zhou2017places}, where P+A means joint training of long-tailed prompts and adaptformers, CC-Init means class-centric initialization, VO and VL mean corresponding model experts.
   }
    \vspace{-1em}
   \label{table:ablation_lptplus}
   \begin{center}
      \setlength{\tabcolsep}{8.6pt} 
	\renewcommand{\arraystretch}{3.2}
	  		{ \fontsize{8.3}{3}\selectfont{
   \begin{tabular}{l|cccccc|cccc}
   \toprule
   {\bf Method}  & \bf LPT & \textbf{P+A} & \textbf{CC-Init} & \textbf{VO} & \textbf{VL} & \textbf{MoE Scorer} &  \bf Overall & \bf Many & {\bf Medium} & {\bf Few}\\ 
   \midrule
   LPT~\cite{dong2023lpt} & \checkmark & - & - & \checkmark & - & - & \textbf{50.1}& 49.3& \textbf{52.3}& \textbf{46.9}\\
   LPT(VL) & \checkmark & - & - & - & \checkmark & - & 50.5 & 51.9 & 53.0 & 42.2  \\
   +Adapter & \checkmark & \checkmark & - & \checkmark & - & - & 50.2 & 47.5 & 53.1 & 48.9  \\
   +Adapter(VL) & \checkmark & \checkmark & - & - & \checkmark & - & 50.8 & 51.3 & 51.5 & 48.9  \\
   LPT++(V) & \checkmark & \checkmark & \checkmark & \checkmark & - & - & 50.5 & \textbf{47.8} & 53.2 & 49.4 \\
   LPT++(VL) & \checkmark & \checkmark & \checkmark & - & \checkmark & - & \textbf{52.2} & 51.7 & \textbf{53.1} & \textbf{51.1}  \\
   +Vanilla Fusion & \checkmark & \checkmark & \checkmark & - & \checkmark & - & \textbf{52.3} & 51.8 & \textbf{53.2} & \textbf{51.3}  \\
   +MoE (LPT++) & \checkmark & \checkmark & \checkmark & \checkmark & \checkmark & \checkmark &\textbf{53.4}& 51.9& \textbf{54.9}& \textbf{52.7} \\
   \bottomrule
   \end{tabular}}}
   \vspace{-2em}
\end{center}
\end{table*}
\begin{table}[h]
\centering
\begin{minipage}[t]{0.48\textwidth}
   \renewcommand\arraystretch{1.3}
   \centering
   \caption{Effect of different pretrained model sizes. 
   }
   \setlength{\tabcolsep}{10.6pt} 
	\renewcommand{\arraystretch}{3.2}
	  		{ \fontsize{8.3}{3}\selectfont{
   \begin{tabular}{l|cc}
      \toprule
      \textbf{Backbone} & \textbf{Phase 1 Acc}  & \textbf{LPT++(V) Acc} \\
      \midrule
      ViT-T & 32.55 & \textbf{37.40} \\
      ViT-S & 40.50 & \textbf{44.66} \\
      ViT-B & 49.41 & \textbf{50.50} \\
      \bottomrule
   \end{tabular}}}
   \label{table:ablation_scale}
\end{minipage}
\vspace{-2em}
\end{table}

\textbf{Results on iNaturalist. } 
We evaluated the performance of LPT on the fine-grained iNaturalist 2018~\cite{van2018inaturalist}, with the results presented in Table~\ref{table:comp_inat18}. LPT achieves an overall accuracy of 76.1\% and a few-shot accuracy of 79.3\%, surpassing all other state-of-the-art methods that utilize vision-only pretrained models. Remarkably, LPT also outperforms the fully fine-tuned ViT-L/16~\cite{touvron2022things} by 0.2\%. These findings demonstrate LPT's capability to effectively manage large-scale, long-tailed datasets through prompt tuning alone while maintaining competitive accuracy. Furthermore, with the integration of the mixture of long-tailed experts, LPT++ achieves an impressive overall accuracy of 82.1\%, outperforming all comparative methods. These results underscore the efficacy of LPT++ in addressing the challenges of large-scale long-tailed learning.


\subsection{Evaluation for Robustness with Domain Shift}\label{sec:full_domain_shift}
Next we investigate the robustness of LPT++ against domain shift.
Since CLIP~\cite{radford2021learning} leverage massive training data from various domains (including ImageNet variants). For fair comparison, we keep using ImageNet-21K pretrained ViT~\cite{dosovitskiy2021an} as backbone (\emph{i.e.}, LPT++(V) and corresponding baselines) to explore the robustness. 
To comprehensively compare LPT++(V) with baseline methods, e.g., linear probe, full fine-tuning, VPT~\cite{jia2022vpt}, and WISE-FT~\cite{wortsman2022robust}, we evaluate these models across six distinct out-of-distribution (OOD) datasets: ImageNet-Sketch~\cite{wang2019learning}, ImageNet-ReaL~\cite{imagenetreal}, ImageNet-V2~\cite{imagenetv2}, ImageNet-A~\cite{imageneta}, ImageNet-R~\cite{imagenetr}, and ObjectNet~\cite{objectnet}. For fairness, all methods were initialized from the same IN21K pretrained ViT-B and fine-tuned on the identical ImageNet-LT training set. The evaluation results, detailed in Table~\ref{table:comp_robustness_full}, demonstrate that LPT++(V) outperforms all baseline methods across all six OOD datasets. Note, as mentioned in \cite{herrmann2022pyramid}, the pretrained ViT~\cite{dosovitskiy2021an} used in our experiments tends to perform suboptimally on the ObjectNet benchmark (e.g., 17.36\% accuracy after ImageNet-1K training). Consequently, the results for ObjectNet reported in the table are relatively modest, reflecting the challenges of training on ImageNet-LT.

%


\subsection{Ablation Study of LPT++}\label{sec:ablation}

We first focus on long-tailed prompt and corresponding training schedule to investigate LPT. Table~\ref{table:ablation} presents the ablation study results for the shared prompt tuning phase and the group-specific prompt tuning phase. In this analysis, we use linear probing~\cite{dosovitskiy2021an} and VPT~\cite{jia2022vpt} as baseline methods. After completing Phase 1 training, both type (a) and type (b) outperform their respective baselines by 8.04\% and 11.58\% in overall accuracy. Additionally, when prompts are introduced for fine-tuning, type (b) shows improvements of 7.77\% in overall accuracy and 15.74\% in few-shot accuracy compared to type (a). These findings suggest that: \textbf{1)} integrating prompts for fine-tuning enhances both overall performance and accuracy for tail classes in long-tailed learning, and \textbf{2)} Phase 1 of LPT effectively leverages the representational capabilities of learnable prompts, leading to superior classification outcomes. Furthermore, replacing cross-entropy loss with $\mathcal{L}_{\text{A-GCL}}$ in type (b) results in type (c) achieving an overall accuracy of 49.41\%, with a 4.07\% improvement in few-shot accuracy. Finally, introducing group-specific prompts and Phase 2 in LPT, type (d) reaches 50.07\% overall accuracy on Places-LT, indicating that using different group prompts for different input samples reduces the complexity of long-tailed learning and further improves classification performance.

Based on the investigation of long-tailed prompts, we further conduct ablation studies to assess the impact of each newly proposed contribution in LPT++, as shown in Table~\ref{table:ablation_lptplus}. After integrating visual adapter~\cite{chen2022adaptformer}, LPT+Adapter maintains the same overall accuracy, but the accuracy for medium-shot and few-shot classes increases to 53.1\% and 48.9\%, respectively. These results demonstrate that advanced parameter-efficient tuning modules benefit tail classes in long-tailed learning, providing a more stable baseline for PEFT-based long-tailed classification. By adopting class-centric initialization, the proposed LPT++(V) achieves 50.5\% overall accuracy and 49.4\% accuracy for few-shot classes. These results indicate that class-centric initialization improves the discriminative ability of long-tailed learners by providing a clearer initial classification boundary, which is particularly beneficial for few-shot classes with limited training samples. Similarly, after applying these strategies, the proposed LPT++(VL) backbone surpasses LPT(VL) by 1.7\% in overall accuracy.

Finally, when using a vanilla fusion approach for both LPT++(V) and LPT++(VL), the accuracy sees only marginal improvement across all classes. However, by adopting our proposed mixture of long-tailed experts with the MoE scorer, the LPT++ achieves a state-of-the-art 53.4\% overall accuracy. 

\begin{table}[t]
   \caption{Comparison of our mixture of long-tailed experts. 
   }
    \vspace{-1em}
   \label{table:comp_fusion}
   \begin{center}
      \setlength{\tabcolsep}{5.6pt} 
	\renewcommand{\arraystretch}{3.2}
	  		{ \fontsize{8.3}{3}\selectfont{
   \begin{tabular}{l|cccc}
   \toprule
   {\bf Method} &  \bf Overall & \bf Many & {\bf Medium} & {\bf Few}\\ 
   \hline
   \multicolumn{5}{l}{\bf Single model} \\
   \hline
   VL-LTR~\cite{tian2021vl} & {50.1}& {54.2}& 48.5& 42.0\\
   LPT++(V)  & 50.5 & {47.8} & 53.2 & 49.4 \\
   LPT++(VL) & {52.2} & 51.7 & {53.1} & {51.1}  \\
   \hline
   \multicolumn{5}{l}{\bf Ensemble models} \\
   \hline
   Vanilla~\cite{sagi2018ensemble} & {52.3} & 51.8 & {53.2} & {51.3}  \\
   Vanilla~\cite{sagi2018ensemble} (3 models) & 52.7 & 52.6 & 53.1 & 52.4  \\
   WISE-FT~\cite{wortsman2022robust} & 52.5 & 51.9 & 53.4 & 51.5  \\
   WISE-FT~\cite{wortsman2022robust} (3 models) & 53.0 & 52.7 & 53.4 & 52.7  \\
   Model Soup~\cite{wortsman2022model} & {52.3} & 51.8 & {53.2} & {51.3} \\
   Model Soup~\cite{wortsman2022model} (3 models) & \textbf{52.7} & 52.6 & {52.9} & {52.5}  \\
   Ours &\textbf{53.4}& 51.9& \textbf{54.9}& \textbf{52.7} \\
   \bottomrule
   \end{tabular}}}
   \vspace{-1em}
\end{center}
\end{table}

\begin{table}[t]
   \caption{Ablation study among different MoE scorers. 
   }
    \vspace{-1em}
   \label{table:comp_moescorer}
   \begin{center}
      \setlength{\tabcolsep}{5.6pt} 
	\renewcommand{\arraystretch}{3.2}
	  		{ \fontsize{8.3}{3}\selectfont{
   \begin{tabular}{cc|cccc}
   \toprule
   {\bf Learning} & {\bf Searching} &  \bf Overall & \bf Many & {\bf Medium} & {\bf Few}\\ 
   \midrule
   \checkmark & - & {53.0}& {51.0}& \textbf{55.0}& 52.3\\
   - & \checkmark  & 53.2 & {51.7} & 55.0 & 52.5 \\
   \checkmark & \checkmark &\textbf{53.4}& \textbf{51.9}& {54.9}& \textbf{52.7} \\
   \bottomrule
   \end{tabular}}}
   \vspace{-2em}
\end{center}
\end{table}

\begin{table}[t]
   \caption{Effect of combining different experts in LPT++.}
    \vspace{-1em}
   \label{table:comp_base_models}
   \begin{center}
      \setlength{\tabcolsep}{5.6pt} 
	\renewcommand{\arraystretch}{3.2}
	  		{ \fontsize{8.3}{3}\selectfont{
   \begin{tabular}{cc|cc|c}
   \toprule
   {\bf Model 1} & \bf Accuracy 1 & {\bf Model 2} & \bf Accuracy 2 &  \bf Overall \\ 
   \midrule
   VL-LTR~\cite{tian2021vl} & 50.1 & LPT++(V)  & 50.5 & 52.6 \\
   VL-LTR~\cite{tian2021vl} & 50.1 & LPT++(VL)  & 52.2 & 52.8 \\
   LPT++(V) & 50.5 & LPT++(VL) & 52.2 &\textbf{53.4} \\
   \bottomrule
   \end{tabular}}}
   \vspace{-1em}
\end{center}
\end{table}

\subsection{Ablation Study of Mixture of Long-tailed Experts}


\subsubsection{Comparison with Other Ensemble Methods. } 
We first compared our approach with traditional ensemble techniques (\emph{e.g.}, vanilla fusion~\cite{sagi2018ensemble}, WISE-FT~\cite{wortsman2022robust}, and model soup~\cite{wortsman2022model}). In addition to ensembling two LPT++ models, we also included a third model, VL-LTR~\cite{tian2021vl}, in the other methods to generate more competitive results. The comparison results, shown in Table~\ref{table:comp_fusion}, indicate that, for ensembles with two base models, all counterparts achieve only marginal improvements in overall accuracy. However, our method achieves a 1.2\% improvement in overall accuracy, along with significant gains of at least 1.6\% in medium-shot and few-shot accuracy. Even when compared to ensembles with three base models, our method with only two base models still achieves a 0.4\% improvement in overall accuracy and a 1.5\% improvement in medium-shot accuracy. These results underscore the effectiveness of our proposed mixture of long-tailed experts in PEFT-based long-tailed learning.


\subsubsection{Which Mixture of Long-tailed Expert is Better? }
Table~\ref{table:comp_moescorer} illustrates the accuracy of LPT++ with different MoE scorers. When using only a learning-based MoE scorer, LPT++ achieves 53.0\% overall accuracy and 51.0\% many-shot accuracy. This suggests that directly learning the weights for the mixture of long-tailed experts may lead to suboptimal solutions. With only a search-based MoE scorer, where all test samples share the same automatically searched MoE weight, the overall accuracy improves to 53.2\%, indicating that this can serve as a feasible initial value for MoE weights. Finally, by combining both methods to learn the offset of MoE weights for input images, LPT++ achieves 53.4\% overall accuracy, indicating that proper weight initialization can ease the difficulty of learning-based MoE scoring for precise predictions.

\begin{table}
   \centering
   \caption{Effect of the hidden dimension of MoE scorer. 
   }
   \vspace{-0.5em}
   \setlength{\tabcolsep}{13.6pt} 
	\renewcommand{\arraystretch}{3.2}
	  		{ \fontsize{8.3}{3}\selectfont{
   \begin{tabular}{l|c|c|c}
      \toprule
      \bf Hidden Dimension & 1024  & 2048 & 4096 \\
      \midrule
      \textbf{Overall Acc} & 53.2 & \textbf{53.4} & \underline{53.4} \\
      \bottomrule
   \end{tabular}}}
   \label{table:hidden_dim}
\vspace{-2em}
\end{table}

\subsubsection{Effects from Different Base Models. }

Mixture of long-tailed experts (MoLEs) is model-agnostic and can be applied to various base model pairs. In addition to using LPT++(V) and LPT++(VL) as model experts, we introduced the CLIP-based long-tailed learning method VL-LTR~\cite{tian2021vl} as another model expert. The results in Table~\ref{table:comp_base_models} reveal two key findings. First, MoLEs improves final performance across different base models, with improvements ranging from 0.6\% to 2.1\%, confirming the model-agnostic nature of our method. Second, combining visual-only and visual-language pretrained models as base models yields better performance. For example, combining CLIP-pretrained LPT++(VL) with VL-LTR results in a 0.6\% increase in overall accuracy. However, when the mixture of long-tailed experts is applied to a pair of visual-only models (e.g., LPT++(V)) and visual-language models (e.g., VL-LTR or LPT++(VL)), the overall accuracy increases by 2.1\% and 1.2\%, respectively. These results indicate that our method fully leverages the strengths of both models to achieve higher accuracy in long-tailed learning.

\subsubsection{Effects of hidden dimension of MoE scorer. } 

Intuitively, a larger MoE scorer might enhance final performance. The corresponding analysis and results are shown in Table~\ref{table:hidden_dim}. One can find that a relatively small hidden dimension (e.g., 1024) performs worse, while larger hidden dimensions tend to achieve higher final accuracy, indicating that a larger MoE scorer can benefit LPT++. However, further increasing the hidden dimension does not lead to additional improvements. 


\section{Conclusion}\label{sec:conclusion}

We present LPT++, a versatile framework for long-tailed classification that combines parameter-efficient fine-tuning with model ensemble. LPT++ enhances frozen ViTs by integrating three key components. First is universal long-tailed adaptation module, which aggregates both long-tailed prompts and visual adapters to adapt the pretrained model to the target domain and improve discriminative ability. Second is mixture of long-tailed experts framework with corresponding MoE scorer, which can adaptively calculate reweight coefficients for confidence scores from visual-only and visual-language models experts to obtain more precise prediction. And finally is the three-phase training framework. By learning each critical module separately, one can obtain a promising long-tailed classification network stably and effectively. 
We also propose LPT, which only incorporates visual-only pretrained ViT alongside the long-tailed prompts by single model based approach. Compared to LPT++, LPT can clearly show the effectiveness of each kind of prompt meanwhile achieving comparable performance without visual-language pretrained models. 
Experimental results on long-tailed benchmarks show that with only ~1\% additional trainable parameters, LPT++ achieves state-of-the-art accuracy, surpassing all counterparts.

{
\bibliographystyle{IEEEtran}
\bibliography{egbib}

\begin{thebibliography}{10}
\providecommand{\url}[1]{#1}
\csname url@samestyle\endcsname
\providecommand{\newblock}{\relax}
\providecommand{\bibinfo}[2]{#2}
\providecommand{\BIBentrySTDinterwordspacing}{\spaceskip=0pt\relax}
\providecommand{\BIBentryALTinterwordstretchfactor}{4}
\providecommand{\BIBentryALTinterwordspacing}{\spaceskip=\fontdimen2\font plus
\BIBentryALTinterwordstretchfactor\fontdimen3\font minus
  \fontdimen4\font\relax}
\providecommand{\BIBforeignlanguage}[2]{{%
\expandafter\ifx\csname l@#1\endcsname\relax
\typeout{** WARNING: IEEEtran.bst: No hyphenation pattern has been}%
\typeout{** loaded for the language `#1'. Using the pattern for}%
\typeout{** the default language instead.}%
\else
\language=\csname l@#1\endcsname
\fi
#2}}
\providecommand{\BIBdecl}{\relax}
\BIBdecl

\bibitem{Kang2020Decoupling}
B.~Kang, S.~Xie, M.~Rohrbach, Z.~Yan, A.~Gordo, J.~Feng, and Y.~Kalantidis,
  ``Decoupling representation and classifier for long-tailed recognition,'' in
  \emph{ICLR}, 2020.

\bibitem{zhang2021deep}
Y.~Zhang, B.~Kang, B.~Hooi, S.~Yan, and J.~Feng, ``Deep long-tailed learning: A
  survey,'' \emph{arXiv preprint arXiv:2110.04596}, 2021.

\bibitem{van2018inaturalist}
G.~Van~Horn, O.~Mac~Aodha, Y.~Song, Y.~Cui, C.~Sun, A.~Shepard, H.~Adam,
  P.~Perona, and S.~Belongie, ``The inaturalist species classification and
  detection dataset,'' in \emph{CVPR}, 2018.

\bibitem{zhou2017places}
B.~Zhou, A.~Lapedriza, A.~Khosla, A.~Oliva, and A.~Torralba, ``Places: A 10
  million image database for scene recognition,'' \emph{IEEE Transactions on
  Pattern Analysis and Machine Intelligence}, 2017.

\bibitem{gupta2019lvis}
A.~Gupta, P.~Dollar, and R.~Girshick, ``Lvis: A dataset for large vocabulary
  instance segmentation,'' in \emph{CVPR}, 2019.

\bibitem{li2019gradient}
B.~Li, Y.~Liu, and X.~Wang, ``Gradient harmonized single-stage detector,'' in
  \emph{Proceedings of the AAAI conference on artificial intelligence},
  vol.~33, no.~01, 2019, pp. 8577--8584.

\bibitem{Li2022Long}
M.~Li, Y.~Cheung, and Y.~Lu, ``Long-tailed visual recognition via gaussian
  clouded logit adjustment,'' in \emph{CVPR}, 2022.

\bibitem{li2021metasaug}
S.~Li, K.~Gong, C.~H. Liu, Y.~Wang, F.~Qiao, and X.~Cheng, ``Metasaug: Meta
  semantic augmentation for long-tailed visual recognition,'' in \emph{CVPR},
  2021.

\bibitem{ren2020metasoftmax}
J.~Ren, C.~Yu, s.~sheng, X.~Ma, H.~Zhao, S.~Yi, and h.~Li, ``Balanced
  meta-softmax for long-tailed visual recognition,'' in \emph{NeurIPS},
  H.~Larochelle, M.~Ranzato, R.~Hadsell, M.~Balcan, and H.~Lin, Eds., 2020.

\bibitem{cui2019cbloss}
\BIBentryALTinterwordspacing
Y.~Cui, M.~Jia, T.-Y. Lin, Y.~Song, and S.~Belongie, ``Class-balanced loss
  based on effective number of samples,'' \emph{CVPR}, 2019. [Online].
  Available: \url{http://dx.doi.org/10.1109/CVPR.2019.00949}
\BIBentrySTDinterwordspacing

\bibitem{menon2021longtail}
A.~K. Menon, S.~Jayasumana, A.~S. Rawat, H.~Jain, A.~Veit, and S.~Kumar,
  ``Long-tail learning via logit adjustment,'' in \emph{ICLR}, 2021.

\bibitem{li2021self}
T.~Li, L.~Wang, and G.~Wu, ``Self supervision to distillation for long-tailed
  visual recognition,'' in \emph{ICCV}, 2021.

\bibitem{zhou2020bbn}
B.~Zhou, Q.~Cui, X.-S. Wei, and Z.-M. Chen, ``Bbn: Bilateral-branch network
  with cumulative learning for long-tailed visual recognition,'' \emph{CVPR},
  2020.

\bibitem{wang2020long}
X.~Wang, L.~Lian, Z.~Miao, Z.~Liu, and S.~X. Yu, ``Long-tailed recognition by
  routing diverse distribution-aware experts,'' \emph{arXiv preprint
  arXiv:2010.01809}, 2020.

\bibitem{cui2021parametric}
J.~Cui, Z.~Zhong, S.~Liu, B.~Yu, and J.~Jia, ``Parametric contrastive
  learning,'' in \emph{ICCV}, 2021.

\bibitem{he2016deep}
K.~He, X.~Zhang, S.~Ren, and J.~Sun, ``Deep residual learning for image
  recognition,'' in \emph{CVPR}, 2016.

\bibitem{dosovitskiy2021an}
A.~Dosovitskiy, L.~Beyer, A.~Kolesnikov, D.~Weissenborn, X.~Zhai,
  T.~Unterthiner, M.~Dehghani, M.~Minderer, G.~Heigold, S.~Gelly, J.~Uszkoreit,
  and N.~Houlsby, ``An image is worth 16x16 words: Transformers for image
  recognition at scale,'' in \emph{ICLR}, 2021.

\bibitem{chen2022adaptformer}
S.~Chen, C.~Ge, Z.~Tong, J.~Wang, Y.~Song, J.~Wang, and P.~Luo, ``Adaptformer:
  Adapting vision transformers for scalable visual recognition,''
  \emph{NeurIPS}, 2022.

\bibitem{li2024uni}
Y.~Li, S.~Jiang, B.~Hu, L.~Wang, W.~Zhong, W.~Luo, L.~Ma, and M.~Zhang,
  ``Uni-moe: Scaling unified multimodal llms with mixture of experts,''
  \emph{arXiv preprint arXiv:2405.11273}, 2024.

\bibitem{dong2023lpt}
B.~Dong, P.~Zhou, S.~Yan, and W.~Zuo, ``{LPT}: Long-tailed prompt tuning for
  image classification,'' in \emph{ICLR}, 2023.

\bibitem{shi2023longtail}
J.-X. Shi, T.~Wei, Z.~Zhou, J.-J. Shao, X.-Y. Han, and Y.-F. Li, ``Long-tail
  learning with foundation model: Heavy fine-tuning hurts,'' 2023.

\bibitem{Jamal_2020_CVPR}
M.~A. Jamal, M.~Brown, M.-H. Yang, L.~Wang, and B.~Gong, ``Rethinking
  class-balanced methods for long-tailed visual recognition from a domain
  adaptation perspective,'' in \emph{CVPR}, 2020.

\bibitem{ma2021simple}
T.~Ma, S.~Geng, M.~Wang, J.~Shao, J.~Lu, H.~Li, P.~Gao, and Y.~Qiao, ``A simple
  long-tailed recognition baseline via vision-language model,'' 2021.

\bibitem{tian2021vl}
C.~Tian, W.~Wang, X.~Zhu, X.~Wang, J.~Dai, and Y.~Qiao, ``Vl-ltr: Learning
  class-wise visual-linguistic representation for long-tailed visual
  recognition,'' \emph{ECCV}, 2022.

\bibitem{Long2022}
A.~Long, W.~Yin, T.~Ajanthan, V.~Nguyen, P.~Purkait, R.~Garg, C.~Shen, and
  A.~van~den Hengel, ``Retrieval augmented classification for long-tail visual
  recognition,'' in \emph{CVPR}, 2022.

\bibitem{lester2021power}
B.~Lester, R.~Al-Rfou, and N.~Constant, ``The power of scale for
  parameter-efficient prompt tuning,'' in \emph{EMNLP}, 2021.

\bibitem{jia2022vpt}
M.~Jia, L.~Tang, B.-C. Chen, C.~Cardie, S.~Belongie, B.~Hariharan, and S.-N.
  Lim, ``Visual prompt tuning,'' in \emph{ECCV}, 2022.

\bibitem{pmlr-v97-houlsby19a}
N.~Houlsby, A.~Giurgiu, S.~Jastrzebski, B.~Morrone, Q.~De~Laroussilhe,
  A.~Gesmundo, M.~Attariyan, and S.~Gelly, ``Parameter-efficient transfer
  learning for {NLP},'' in \emph{ICML}, 2019.

\bibitem{he2022towards}
J.~He, C.~Zhou, X.~Ma, T.~Berg-Kirkpatrick, and G.~Neubig, ``Towards a unified
  view of parameter-efficient transfer learning,'' in \emph{ICLR}, 2022.

\bibitem{hu2022lora}
E.~J. Hu, yelong shen, P.~Wallis, Z.~Allen-Zhu, Y.~Li, S.~Wang, L.~Wang, and
  W.~Chen, ``Lo{RA}: Low-rank adaptation of large language models,'' in
  \emph{ICLR}, 2022.

\bibitem{zhai2019largescale}
X.~Zhai, J.~Puigcerver, A.~Kolesnikov, P.~Ruyssen, C.~Riquelme, M.~Lucic,
  J.~Djolonga, A.~S. Pinto, M.~Neumann, A.~Dosovitskiy, L.~Beyer, O.~Bachem,
  M.~Tschannen, M.~Michalski, O.~Bousquet, S.~Gelly, and N.~Houlsby, ``A
  large-scale study of representation learning with the visual task adaptation
  benchmark,'' 2019.

\bibitem{zhou2017scene}
B.~Zhou, H.~Zhao, X.~Puig, S.~Fidler, A.~Barriuso, and A.~Torralba, ``Scene
  parsing through ade20k dataset,'' in \emph{CVPR}, 2017.

\bibitem{deng2009imagenet}
J.~Deng, W.~Dong, R.~Socher, L.-J. Li, K.~Li, and L.~Fei-Fei, ``Imagenet: A
  large-scale hierarchical image database,'' in \emph{CVPR}, 2009.

\bibitem{wang2022learning}
Z.~Wang, Z.~Zhang, C.-Y. Lee, H.~Zhang, R.~Sun, X.~Ren, G.~Su, V.~Perot, J.~Dy,
  and T.~Pfister, ``Learning to prompt for continual learning,'' in
  \emph{CVPR}, 2022.

\bibitem{zhou2022learning}
K.~Zhou, J.~Yang, C.~C. Loy, and Z.~Liu, ``Learning to prompt for
  vision-language models,'' \emph{International Journal of Computer Vision},
  2022.

\bibitem{balakrishnama1998linear}
S.~Balakrishnama and A.~Ganapathiraju, ``Linear discriminant analysis-a brief
  tutorial,'' \emph{Institute for Signal and information Processing}, vol.~18,
  no. 1998, pp. 1--8, 1998.

\bibitem{mackiewicz1993principal}
A.~Ma{\'c}kiewicz and W.~Ratajczak, ``Principal components analysis (pca),''
  \emph{Computers \& Geosciences}, vol.~19, no.~3, pp. 303--342, 1993.

\bibitem{hinton2002stochastic}
G.~E. Hinton and S.~Roweis, ``Stochastic neighbor embedding,'' \emph{NeurIPS},
  2002.

\bibitem{vaswani2017attention}
A.~Vaswani, N.~Shazeer, N.~Parmar, J.~Uszkoreit, L.~Jones, A.~N. Gomez,
  {\L}.~Kaiser, and I.~Polosukhin, ``Attention is all you need,'' in
  \emph{NeurIPS}, 2017.

\bibitem{radford2021learning}
A.~Radford, J.~W. Kim, C.~Hallacy, A.~Ramesh, G.~Goh, S.~Agarwal, G.~Sastry,
  A.~Askell, P.~Mishkin, J.~Clark \emph{et~al.}, ``Learning transferable visual
  models from natural language supervision,'' in \emph{ICML}, 2021.

\bibitem{openai2023gpt4}
OpenAI, ``Gpt-4 technical report,'' 2023.

\bibitem{ridnik2021asymmetric}
T.~Ridnik, E.~Ben-Baruch, N.~Zamir, A.~Noy, I.~Friedman, M.~Protter, and
  L.~Zelnik-Manor, ``Asymmetric loss for multi-label classification,'' in
  \emph{ICCV}, 2021.

\bibitem{lin2017focal}
T.-Y. Lin, P.~Goyal, R.~Girshick, K.~He, and P.~Doll{\'a}r, ``Focal loss for
  dense object detection,'' in \emph{ICCV}, 2017.

\bibitem{openlongtailrecognition}
Z.~Liu, Z.~Miao, X.~Zhan, J.~Wang, B.~Gong, and S.~X. Yu, ``Large-scale
  long-tailed recognition in an open world,'' in \emph{CVPR}, 2019.

\bibitem{wang2019learning}
H.~Wang, S.~Ge, Z.~Lipton, and E.~P. Xing, ``Learning robust global
  representations by penalizing local predictive power,'' in \emph{NeurIPS},
  2019.

\bibitem{imageneta}
D.~Hendrycks, K.~Zhao, S.~Basart, J.~Steinhardt, and D.~Song, ``Natural
  adversarial examples,'' \emph{CVPR}, 2021.

\bibitem{imagenetr}
D.~Hendrycks, S.~Basart, N.~Mu, S.~Kadavath, F.~Wang, E.~Dorundo, R.~Desai,
  T.~Zhu, S.~Parajuli, M.~Guo, D.~Song, J.~Steinhardt, and J.~Gilmer, ``The
  many faces of robustness: A critical analysis of out-of-distribution
  generalization,'' \emph{ICCV}, 2021.

\bibitem{imagenetv2}
B.~Recht, R.~Roelofs, L.~Schmidt, and V.~Shankar, ``Do imagenet classifiers
  generalize to imagenet?'' in \emph{ICML}, 2019.

\bibitem{touvron2022things}
H.~Touvron, M.~Cord, A.~El-Nouby, J.~Verbeek, and H.~Jégou, ``Three things
  everyone should know about vision transformers,'' 2022.

\bibitem{wortsman2022robust}
M.~Wortsman, G.~Ilharco, J.~W. Kim, M.~Li, S.~Kornblith, R.~Roelofs, R.~G.
  Lopes, H.~Hajishirzi, A.~Farhadi, H.~Namkoong \emph{et~al.}, ``Robust
  fine-tuning of zero-shot models,'' in \emph{CVPR}, 2022.

\bibitem{imagenetreal}
L.~Beyer, O.~J. H{\'e}naff, A.~Kolesnikov, X.~Zhai, and A.~v.~d. Oord, ``Are we
  done with imagenet?'' \emph{arXiv preprint arXiv:2006.07159}, 2020.

\bibitem{objectnet}
A.~Barbu, D.~Mayo, J.~Alverio, W.~Luo, C.~Wang, D.~Gutfreund, J.~Tenenbaum, and
  B.~Katz, ``Objectnet: A large-scale bias-controlled dataset for pushing the
  limits of object recognition models,'' in \emph{NeurIPS}, H.~Wallach,
  H.~Larochelle, A.~Beygelzimer, F.~d\textquotesingle Alch\'{e}-Buc, E.~Fox,
  and R.~Garnett, Eds.\hskip 1em plus 0.5em minus 0.4em\relax Curran
  Associates, Inc., 2019.

\bibitem{herrmann2022pyramid}
C.~Herrmann, K.~Sargent, L.~Jiang, R.~Zabih, H.~Chang, C.~Liu, D.~Krishnan, and
  D.~Sun, ``Pyramid adversarial training improves vit performance,'' in
  \emph{CVPR}, 2022.

\bibitem{sagi2018ensemble}
O.~Sagi and L.~Rokach, ``Ensemble learning: A survey,'' \emph{Wiley
  interdisciplinary reviews: data mining and knowledge discovery}, vol.~8,
  no.~4, p. e1249, 2018.

\bibitem{wortsman2022model}
M.~Wortsman, G.~Ilharco, S.~Y. Gadre, R.~Roelofs, R.~Gontijo-Lopes, A.~S.
  Morcos, H.~Namkoong, A.~Farhadi, Y.~Carmon, S.~Kornblith \emph{et~al.},
  ``Model soups: averaging weights of multiple fine-tuned models improves
  accuracy without increasing inference time,'' in \emph{ICML}, 2022.

\end{thebibliography}
}


\ifCLASSOPTIONcaptionsoff
  \newpage
\fi

\end{document}